\documentclass{article}

\usepackage[preprint]{neurips_2026}
\usepackage{amsmath,amssymb}
\usepackage{algorithm}
\usepackage{algpseudocode}

\usepackage[utf8]{inputenc} 
\usepackage[T1]{fontenc}    
\usepackage{hyperref}       
\usepackage{url}            
\usepackage{booktabs}       
\usepackage{amsfonts}       
\usepackage{nicefrac}       
\usepackage{microtype}      
\usepackage{xcolor}         
\usepackage{graphicx}       
\usepackage{listings}       
\usepackage[most]{tcolorbox}
\definecolor{promptbg}{HTML}{F6F7F9}
\definecolor{promptframe}{HTML}{3B6EA5}
\lstdefinestyle{promptlst}{
  basicstyle=\scriptsize\ttfamily,breaklines=true,
  breakatwhitespace=false,columns=fullflexible,keepspaces=true,
  showstringspaces=false,aboveskip=0pt,belowskip=0pt,upquote=true}
\newtcblisting[auto counter,number within=section]{verbprompt}[2][]{%
  breakable,enhanced,listing only,listing style=promptlst,
  colback=promptbg,colframe=promptframe,coltitle=white,
  fonttitle=\bfseries\small,boxrule=0.6pt,arc=2pt,
  left=6pt,right=6pt,top=5pt,bottom=5pt,
  title={Prompt \thetcbcounter: #2},#1}

\title{Formatting Instructions For NeurIPS 2026}

\author{
  \textbf{Aditya Kumaran$^{1}$,\quad
  Rahul Singhal$^{1}$,\quad
  Karime Maamari$^{1}$}\\
  \textbf{Amine Mhedhbi$^{2,3}$,\quad
  Pradyumna Tambwekar$^{1}$}\\[2mm]
  \normalfont
  $^{1}$Distyl AI,\quad
  $^{2}$Polytechnique Montr\'eal,\quad
  $^{3}$Mila -- Quebec AI Institute\\[1mm]
  \nolinkurl{{aditya.kumaran,rahul,karime}@distyl.ai}\\
  \nolinkurl{amine.mhedhbi@polymtl.ca};\quad
  \nolinkurl{pradyumna.tambwekar@distyl.ai}
}

\begin{document}

\title{\mbox{\raisebox{-0.35\height}{%
  \includegraphics[height=1em]{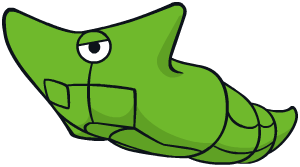}}%
  \hspace{0.18em}HARDEN:} Constrained Evolutionary Search for Harder,
  Answer-Preserving Evaluation Cases}

\maketitle

\begin{abstract}
Language models are often evaluated on curated benchmarks that underrepresent the complexity of enterprise deployments. 
We introduce \textsc{HARDEN},
a constrained evolutionary search method to adapt the input of existing evaluation cases into more challenging variants while keeping their expected outputs fixed. 
\textsc{HARDEN} searches along generated domain-specific complexity axes while enforcing feasibility constraints such as preserving task semantics, realism, and execution validity. 
Across FinQA, PubMedQA, and ContractNLI and three Qwen3.5 model scales (35B-A3B, 122B-A10B, and 397B-A17B), \textsc{HARDEN} reduces task-model accuracy by 22.7\% on average and by up to 49.9\% relative to single-pass baselines using the same feasibility checks. 
These results show that evolutionary search can produce substantially harder valid evaluation cases.
\end{abstract}

\section{Introduction}

Language models (LMs) are increasingly evaluated on curated or synthetic benchmarks~\cite{chen-etal-2021-finqa, guha2023legalbench,jin-etal-2019-pubmedqa, koreeda-manning-2021-contractnli, li2024can}. 
While these benchmarks provide controlled and reproducible evaluations, their inputs often abstract away the ambiguity, grounding requirements, and contextual complexity of real-world domains.
As a result, 
clean benchmarks provide a weak signal of reliability, and high effectiveness on them may not carry over to deployments. We tackle the problem of automatically adapting existing evaluation cases into harder variants for real-world AI systems.

Making a benchmark more challenging, however, is insufficient.
A task modification can reduce model effectiveness by introducing contradictions or removing necessary evidence. 
For example, changing a referenced fiscal period in a financial reasoning case may lower model effectiveness while also changing the correct answer, making the original evaluation target invalid. 
As suggested by Goodhart's law, optimizing a measure can encourage solutions that exploit the measure rather than pursue the intended objective.
Likewise, LM-based case generators optimized only for model failure may increase difficulty by violating task semantics rather than exposing meaningful capability weaknesses. 
We therefore ask: \emph{how can we increase task difficulty (measured by both model performance and model uncertainty) while preserving the evaluation target?}

We formulate this problem as a constrained search for harder evaluation cases. For each search, we start from an existing case whose task and expected output provide a fixed reference. We modify only its input while keeping the expected output fixed.

We introduce \textsc{HARDEN}, an evolutionary search algorithm that modifies existing benchmarks to enhance task complexity and model uncertainty.
\textsc{HARDEN} iteratively proposes and selects increasingly challenging cases while enforcing three constraints:  
(i)~\emph{correctness}, requiring that each variant preserves the evidence needed to derive the expected output; 
(ii)~\emph{realism}, requiring that it remain plausible in the deployment domain; and 
(iii)~\emph{validity}, requiring that it remain well-formed and executable by the original benchmark.

Our contributions are as follows:
\begin{itemize}
    \item \textsc{HARDEN}, a constrained evolutionary search framework that
    adapts existing evaluation cases along domain-specific complexity axes.
    \item LM-based feasibility checks for correctness and realism that discard variants that fail either check, evaluated against human judgments.
    \item An evaluation across professional domains and model sizes shows that, relative to single-pass baselines using the same feasibility checks, \textsc{HARDEN} produces valid cases that \emph{reduce task-model accuracy by 22.7\% on average and up to 49.9\%}, while increasing normalized output semantic entropy by 112\% on average.
\end{itemize}

\section{Related Work}

Prior work generates harder evaluation cases while attempting to preserve their original semantics. Answer-preserving adversarial testing adds distractors to SQuAD without changing the correct answer~\citep{rajpurkar-etal-2016-squad,DBLP:conf/emnlp/JiaL17}, while SEARs derive semantically equivalent replacement rules across multiple NLP tasks~\citep{ribeiro-etal-2018-semantically}. Population-based search has similarly been used to find semantically and syntactically similar adversarial text, with TextAttack later formalizing attacks through transformations, constraints, objectives, and search~\citep{alzantot-etal-2018-generating,morris-etal-2020-textattack}.

More recent methods emphasize constrained and adaptive generation: SECA preserves semantic equivalence and coherence during adversarial prompt search~\citep{NEURIPS2025_d077bc9e}; CETBench and SQLMorph apply semantics-preserving transformations to code and Text-to-SQL evaluation~\citep{oza-etal-2026-llms,DBLP:conf/icde/MalekpourRLM26}; DARG adaptively modifies reasoning graphs while validating labels~\citep{NEURIPS2024_f5198bc2}; and AdvPrompter generates human-readable adversarial suffixes that preserve instruction meaning~\citep{pmlr-v267-paulus25a}.

Unlike prior methods, \textsc{HARDEN} 
treats the correctness, realism, and validity constraints as separate without relying only on semantic similarity or predefined transformations.

\section{Our Approach: \textsc{HARDEN}}
\label{sec:methodology}

We propose a constrained evolutionary method that modifies inputs while holding their evaluation targets fixed.
Let $(x,y)$ denote an original evaluation case as an input-output pair.
The \textsc{HARDEN} search seeks to find a candidate input $x'$ that minimizes the fitness function $s(x')$, defined as follows:
\begin{equation}
\begin{aligned}
x^{\star} \;\in\; \operatorname*{arg\,min}_{x' \in \mathcal{X}(x)} \quad
& s(x') \;=\; \frac{1}{T}\sum_{t=1}^{T}
   \operatorname{Grade}\!\left(\hat{y}_t,\, y\right) \\
\text{s.t.} \quad
& C(x,x') = 1, \quad R(x') = 1, \quad V(x') = 1, \quad s(x') > 0 .
\end{aligned}
\label{eq:fitness-constraints}
\end{equation}
Here, $\mathcal{X}(x)$ denotes the search space of candidate inputs derived from $x$ through possible transformations $\mathcal{X}$. 
The fitness score $s(x')$ estimates the model's performance using the average benchmark score over $T$ sampled responses $\hat{y}_t$, where 
$\operatorname{Grade}$ is the benchmark evaluation function.
The search is subject to three binary feasibility checks:
(i) $C(x,x')$, \emph{correctness}: $x'$ preserves the evidence needed to derive the expected output $y$;
(ii) $R(x')$, \emph{realism}: $x'$ remains plausible in the deployment domain;
(iii) $V(x')$, \emph{artifact validity}: $x'$ remains well-formed and executable by the original benchmark; and an additional score requirement: $s(x')>0$, which excludes cases that collapse the task entirely.

\textbf{Search Domain Analysis.} 
Before the evolutionary search, \textsc{HARDEN} samples the evaluation cases that it will evolve and derives two components used throughout the search: a set of complexity axes and a realism rubric.
First, an agent analyzes the sampled target cases to identify eight to eleven domain-specific \emph{complexity axes} and assign each an initial priority. 
A complexity axis corresponds to a directional strategy that can be applied to the input to make it more difficult for a model to derive the output. 
Second, a web-search agent uses the benchmark details, including its description, schemas, and instructions, together with web sources outside the target benchmark, to construct a rubric for judging whether an input is realistic for the domain. 
The rubric is calibrated on a held-out set before advancement. 
Examples of the realism rubric and complexity axes are provided in Appendices~\ref{app:rubric-example} and~\ref{app:axis-example}, respectively.

\textbf{Evolutionary Search.} 
\textsc{HARDEN} performs an independent search for each evaluation case selected for evolution.
For each case, the original input $x$ initializes the search, which proceeds for $G$ generations, generating $\lambda$ variants at each generation. 
The search uses two distinct and fixed models throughout the search: 
a \emph{mutation model} to generate input variants and a \emph{task model} being evaluated on the variants through the fitness function in Eq.~\ref{eq:fitness-constraints}.

At generation $g \in \{1,\ldots,G\}$, let $x_g^\mathrm{p}$ denote the current parent input, where $x_1^\mathrm{p} = x$.
To generate the $j$-th variant, where $j\in\{1,\ldots,\lambda\}$, \textsc{HARDEN} first selects a complexity axis $a_{g,j}$ from the fixed set of complexity axes $\mathcal{A}$ identified during domain analysis and then uses it to guide the mutation:
\begin{equation}
x'_{g,j}
=
\operatorname{Mutate}
\left(x_g^{\mathrm{p}}; a_{g,j}\right),
\qquad a_{g,j}\in\mathcal{A}.
\label{eq:mutation}
\end{equation}
The selected axis specifies the type of complexity to introduce, while the mutation model determines how to realize it. 
Appendix~\ref{app:mutation-template} provides the mutation prompt.

Axis selection adapts across generations. 
Before each generation, \textsc{HARDEN} ranks the axes using their initial priority from domain analysis,
whether previous applications passed the feasibility constraints and lowered fitness, 
and whether an axis has already been used for the current case. 
It favors high-ranked axes while encouraging different axes across the generated variants.

Each generated variant is then evaluated against the feasibility checks: correctness, realism, and validity in Eq.~\ref{eq:fitness-constraints}. 
The correctness and realism checks are LM-based: each uses $K$ independent model calls and passes when a majority accept the variant. 
Candidates that fail any check are discarded. 
Among the feasible candidates, \textsc{HARDEN} retains only a provided $\mu$ from the lowest fitness scores.
Ties are broken using normalized discrete semantic entropy over sampled outputs~\citep{farquhar2024detecting}, favoring greater model uncertainty, with any remaining tie favoring a newly generated variant.
The best retained candidate is set as $x_{g+1}^{\mathrm{p}}$. 
After $G$ generations, the best candidate $x^\star$ is returned only if $s(x^\star)<s(x)$ and feasibility checks are satisfied; otherwise, the original input is retained.
Appendix~\ref{app:harden-example} shows an original evaluation case and its \textsc{HARDEN}-generated variant.

\section{Experiments}
\label{sec:experiments}

We evaluate three research questions (RQs):
\begin{description}
    \item[\emph{RQ1.}] \emph{Does evolutionary search improve over single-pass baselines using the same feasibility checks enough to justify its additional cost?}
    \item[\emph{RQ2.}] \emph{Does \textsc{HARDEN} remain effective as task-model capability increases?}
    \item[\emph{RQ3.}] \emph{How do \textsc{HARDEN}'s correctness and realism checks compare with human judgment?}
\end{description}



\subsection{Setup}
\label{sec:datasets}

\textbf{Models.} We use \texttt{Qwen3.5} MoEs (35B-A3B, 122B-A10B, and
397B-A17B) as task models served through
OpenRouter with reasoning disabled. We use \texttt{Claude} \texttt{Opus 4.8} for domain analysis and mutation at high reasoning
effort. 



\textbf{Benchmarks.} 
We use FinQA~\citep{chen-etal-2021-finqa},
PubMedQA~\citep{jin-etal-2019-pubmedqa}, and
ContractNLI~\citep{koreeda-manning-2021-contractnli} (part of the
LegalBench suite~\citep{guha2023legalbench}) to represent three different professional domains where deployed solutions 
must handle messy inputs. 
We selected these as near-saturation tasks for the three 
task-model scales \citep{phogat2023zeroshot,singhal2023clinical,nori2023medprompt,schuster2022stretching} (\ref{app:score-agreement}). 
We sample 200 examples as a representative set for each benchmark. 


\textbf{Baselines.} 
We compare against two 
baselines 
representing 
LM-based alternatives. 
\emph{Few-Shot} makes one tool-free mutation per evaluation case.
\emph{Few-Shot (web-search)} makes the same mutation with read-only web access,
while blocking searches that identify the target benchmark. 
Each benchmark case is
paired with five representative challenging evaluation cases sampled from low-scoring examples from a disjoint bank, 
identified as real, difficult few-shot examples.
Appendix~\ref{app:baseline-details} describes how these examples are selected. The baselines use the same models and feasibility checks as \textsc{HARDEN}.

\textbf{\textsc{HARDEN} parameters.}
We run \textsc{HARDEN} with
$G=4$, 
$\lambda=11$, 
$\mu=5$, $K=3$, 
and $T=10$. 

\textbf{Metrics.} 
We use 
each benchmark's evaluation metric: execution
accuracy for FinQA 
and exact label match for
PubMedQA and ContractNLI. We refer to both as \emph{accuracy} for brevity. 
We utilize \emph{semantic entropy}~\citep{farquhar2024detecting,kuhn2023semantic} to measure a model's predictive uncertainty. This metric computes frequency-based uncertainty over the output-distribution of sampled task model outputs. Calculation of additional post-hoc uncertainty metrics is detailed in Appendix~\ref{app:runtime-details}.

\subsection{Results}
\label{sec:results}


\begin{figure}[t]
\centering
\includegraphics[width=\linewidth]{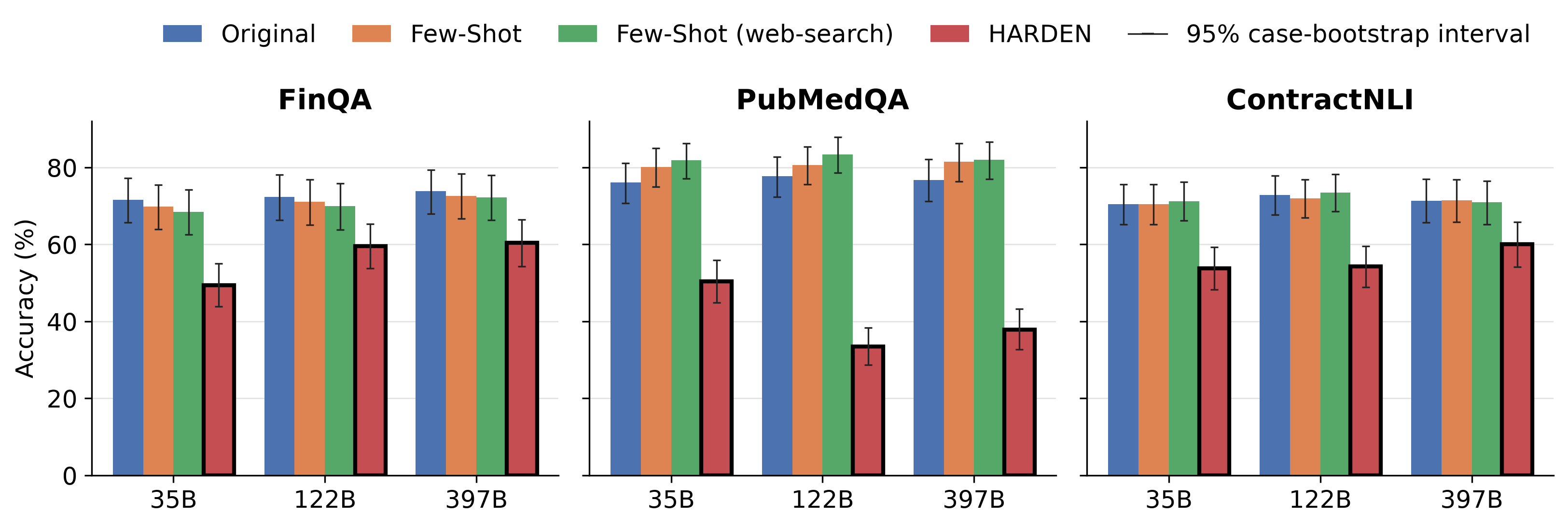}
\caption{Task model accuracy by mutation method and model size. Lower accuracy indicates more effective hardening. Error bars are 95\% case-bootstrap CIs~\citep{efron1993bootstrap} based on ten responses per case.}
\vspace{-1em}
\label{fig:main-accuracy}
\end{figure}

\begin{figure*}[t]
\centering
\includegraphics[width=\linewidth]{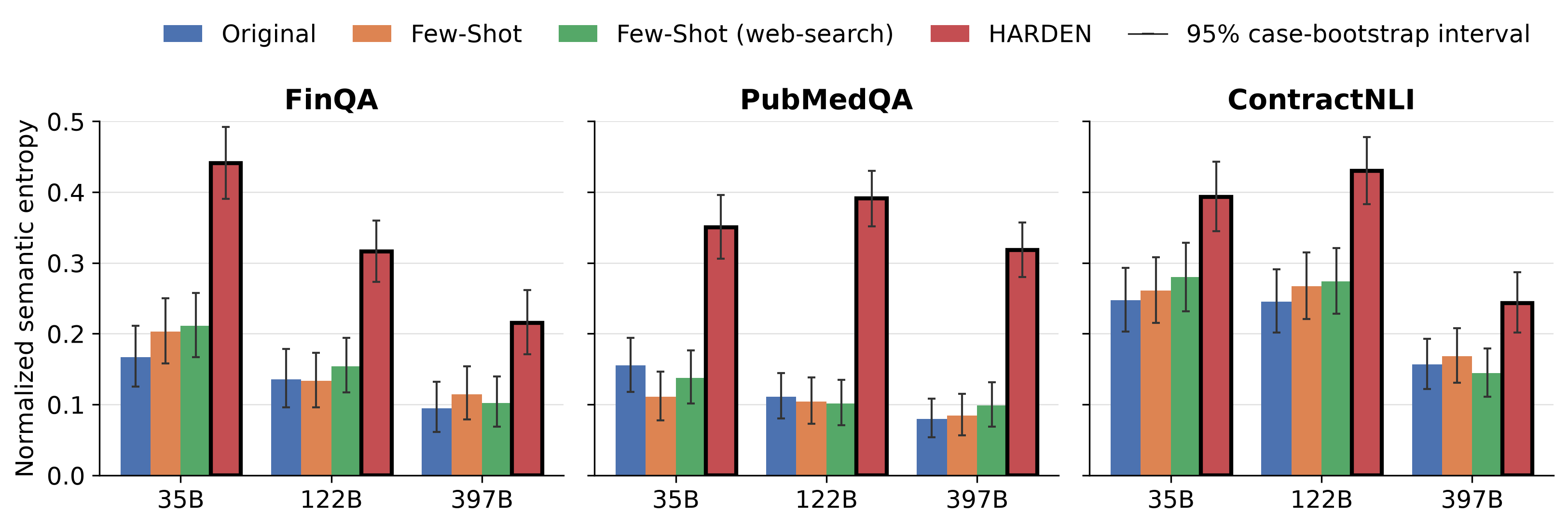}
\caption{Mean-normalized discrete semantic entropy by mutation method and task
model size. Higher indicates greater task-model output uncertainty. Error bars show 95\% case-bootstrap confidence intervals.}
\label{fig:main-uncertainty}
\end{figure*}

\begin{figure*}[t]
\centering
\includegraphics[width=\linewidth]{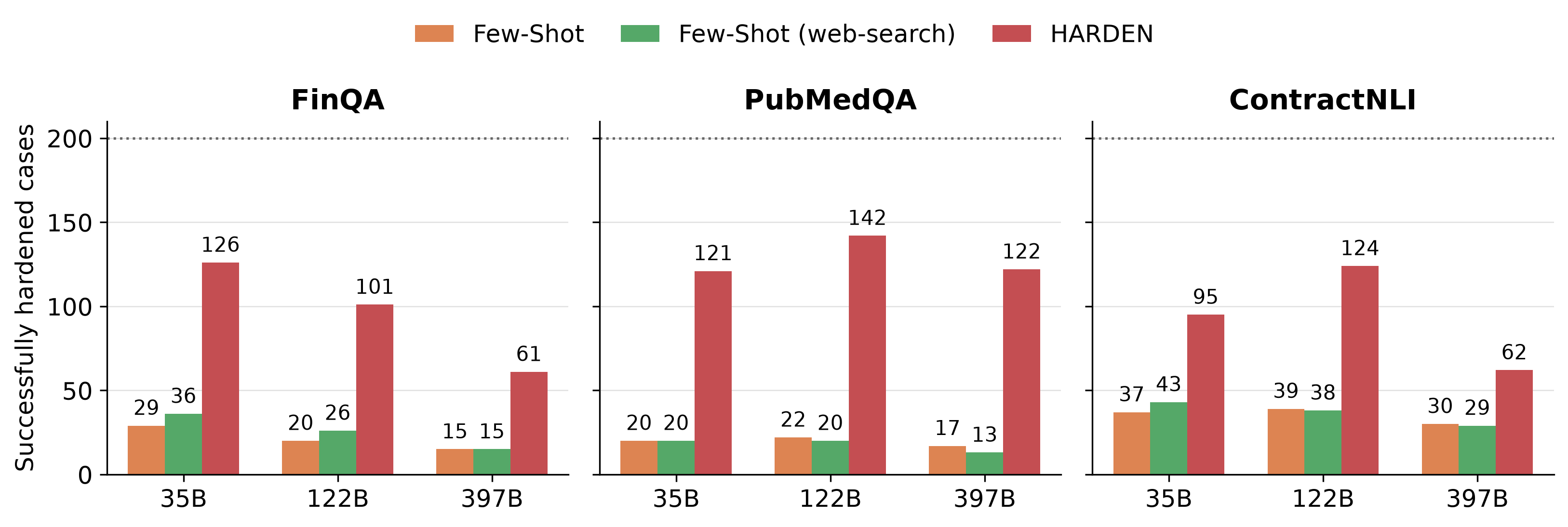}
\caption{Successfully hardened cases (out of 200) by mutation method and
task-model size. 
A case is counted when it passes both LM-based checks and
reduces sampled task-model accuracy relative to its original.}
\label{fig:successful-hardening-counts}
\end{figure*}

Figure~\ref{fig:main-accuracy} reports task-model accuracy over all 200 cases in each of FinQA, PubMedQA, and ContractNLI for the original data and for the datasets produced by both baselines and \textsc{HARDEN}. Figure~\ref{fig:main-uncertainty} reports uncertainty (semantic entropy) for the same cases. Figure~\ref{fig:successful-hardening-counts} reports the number of successfully hardened cases for each method.
Appendix \ref{app:uncertainty-results} reports accuracy and uncertainty metrics, including additional post-hoc uncertainty metrics.

\paragraph{RQ1. \textsc{HARDEN} consistently outperforms single-pass baselines.}
Across all benchmark-model combinations, \textsc{HARDEN} reduces accuracy for $52.4\%$ of the original cases, 
compared with $12.7\%$ for Few-Shot and $13.3\%$ for Few-Shot (web-search). On average, for FinQA, PubMedQA, and ContractNLI, \textsc{HARDEN} reduces the accuracy 
by $16.1\%$, $36.3\%$, and $15.6\%$, respectively, across all three Qwen task models. 
In contrast, Few-Shot changes the accuracy by $-1.4\%$ (FinQA), $+3.9\%$ (PubMedQA), and
$-0.3\%$ (ContractNLI) across the three domains, while Few-Shot (web-search) changes it by
$-2.4\%$, $+5.6\%$, and $+0.3\%$. 
The correctness and realism checks discard $42.0\%$, $3.5\%$, and $23.0\%$ of Few-Shot mutations on FinQA, PubMedQA, and ContractNLI, respectively, and $27.0\%$, $2.0\%$, and $22.5\%$ of Few-Shot (web-search) mutations. Semantic entropy shows the same separation: averaged across model scales, \textsc{HARDEN} increases entropy by 0.19 on FinQA, 0.24 on PubMedQA, and 0.13 on ContractNLI, while both single-pass baselines remain within 0.02 of the original average in each domain. \textsc{HARDEN} consistently identifies feasible mutations that yield lower task-model accuracy and higher output uncertainty, indicating greater predictive difficulty across all three domains.
In comparison, single-pass mutation, even with web access, rarely achieves both objectives of  feasibility (correctness and realism) and increased difficulty. 

\paragraph{RQ2. \textsc{HARDEN} remains effective as task-model size increases.}
Across the three benchmarks, \textsc{HARDEN} reduces accuracy by an average of $21.5\%$, $25.2\%$, and $21.2\%$ for the 35B, 122B, and 397B Qwen task models, respectively. 
The reduction remains substantial and consistent as model size increases. At the largest model size, HARDEN successfully hardens an average of 81.7 out of 200 cases for a benchmark, nearly 2.5 times the 33.0 achieved by the strongest baseline at the smallest model size (Figure~\ref{fig:successful-hardening-counts}). Across the three benchmarks, \textsc{HARDEN} also increases semantic entropy from 0.19 to 0.39 at 35B, 0.17 to 0.38 at 122B, and 0.11 to 0.26 at 397B, averaging 112\% above the constraint-filtered single-pass baselines (Figure~\ref{fig:main-uncertainty}). The baseline approaches produce minimal shifts in task performance or predictive uncertainty. These results show that \textsc{HARDEN} continues to identify and introduce challenging complexity as task-model size and capability grow. 

\paragraph{RQ3. Agreement of feasibility checks with human judgments.}
We compare \textsc{HARDEN}'s correctness and realism checks with human judgments on the same cases (\ref{app:check-validation}).
For realism, human participants and \textsc{HARDEN} independently evaluate 60 cases per benchmark: 20 original cases and 40 synthetic cases, comprising 20 generated by \textsc{GPT-5.6 Sol} and 20 by \textsc{GPT-3.5}, representing stronger and weaker generators, respectively.
For correctness, they evaluate 20 original--mutated pairs from \textsc{HARDEN} runs for each of FinQA and PubMedQA, balanced between 10 correct and 10 incorrect mutations as measured by the auditor. In both studies, the human label is two-of-three majority vote.

On FinQA, the realism check accepts $90\%$, $95\%$, and $0\%$ of the original, \textsc{GPT-5.6 Sol}-generated, and \textsc{GPT-3.5}-generated cases, versus $60\%$, $70\%$, and $60\%$ for participants.
Direct agreement is $51.7\%$.
PubMedQA shows directional alignment and $85.0\%$ agreement, although the check accepts $90\%$ of \textsc{GPT-3.5}-generated cases. 
Thus, \textsc{HARDEN} is more selective than participants when filtering weaker-generator variants on FinQA, and more closely tracks their judgments on PubMedQA. 

For correctness, across the 40 mutations from both benchmarks, participant majorities accept 39, including 19 of the 20 rejected by the check, providing little discrimination.
To distinguish indiscriminate rejection from conservative handling of ambiguous cases, we evaluated the check on a separate FinQA set specifically to test whether it could identify incorrect examples. We generated counterfactuals by altering critical evidence to invalidate the original output, included correctness-preserving controls, and retained only mutations with agreed co-author labels; the check achieved 85.7\% agreement. Overall, the study shows greater participant agreement for realism on PubMedQA than on FinQA. The correctness check appears to identify incorrect examples, albeit conservatively, but
participant judgments provide limited evidence for validation.


\section{Limitations}
\label{sec:limitations}

The experiments cover benchmarks representing three professional domains, and use one task model
family and one mutation model.  Trends across Qwen3.5 model scales do not establish transfer across architectures.
All task models are evaluated with reasoning disabled, so whether reasoning changes robustness to the mutated examples remains open.
Cross-model transfer of difficulty remains to be measured.  The feasibility check validation studies are small
and provide mixed evidence.  Participants were role-screened, but their domain expertise was not additionally verified as a part of the studies. The initial correctness study provides limited evidence about the check’s ability to distinguish correct from incorrect mutations, and the
secondary adjudicated study covers only 21 FinQA pairs.  Additional evidence is required for conclusive proof of validity.
Addressing these limitations requires larger studies, a redesigned participant correctness interface, evaluation of naturally occurring invalid mutations in every domain, and recalibration of check thresholds
against those judgments. Finally, our approach is substantially more expensive than single-pass mutation, due to evolutionary expansion of mutated case candidates across multiple generations. We encourage practitioners to gauge whether the benefits provided by the approach are worthwhile prior to leveraging the approach.

\section{Conclusion}
\label{sec:conclusion}

Curated benchmarks used for the evaluation of AI systems often fail to emulate the complexity and ambiguity of real-world data. We introduce \textsc{HARDEN}, a constrained evolutionary search method to inject domain-realistic
complexity into benchmarks while preserving their evaluation targets.
Across FinQA, PubMedQA, and ContractNLI, \textsc{HARDEN} significantly increases task difficulty, as measured by accuracy and uncertainty, relative to single-pass agentic baselines.  
The effect remains substantial as
Qwen3.5 task model capability increases from 35B to 122B to 397B, showing that
model-conditioned evolutionary search retains efficacy as task models strengthen. These results establish \textsc{HARDEN} as an effective method for constructing harder, answer-preserving evaluations across the studied domains and model scales. 

\begin{ack}
We thank our colleagues Benjamin Fowlersmith, Deep Patel, Durga Sandeep Saluru, and Philippe Wyder at Distyl AI for their contributions to the greater system from which the design for \textsc{HARDEN} was inspired. The Metapod artwork in the title is sourced from PNGKey~\citep{pngkeyMetapod} under the site’s stated noncommercial-use terms. The original artist is not identified on the source page.
\end{ack}

\bibliographystyle{abbrv}
\bibliography{references}

@article{li2024can,
  title={Can llm already serve as a database interface? a big bench for large-scale database grounded text-to-sqls},
  author={Li, Jinyang and Hui, Binyuan and Qu, Ge and Yang, Jiaxi and Li, Binhua and Li, Bowen and Wang, Bailin and Qin, Bowen and Geng, Ruiying and Huo, Nan and others},
  journal={Advances in Neural Information Processing Systems},
  volume={36},
  year={2024}
}

@inproceedings{DBLP:conf/emnlp/JiaL17,
  author       = {Robin Jia and
                  Percy Liang},
  title        = {Adversarial Examples for Evaluating Reading Comprehension Systems},
  booktitle    = {EMNLP},
  pages        = {2021--2031},
  year         = {2017},
}

@inproceedings{ribeiro-etal-2018-semantically,
    title = "Semantically Equivalent Adversarial Rules for Debugging {NLP} models",
    author = "Ribeiro, Marco Tulio  and
      Singh, Sameer  and
      Guestrin, Carlos",
    booktitle = {ACL},
    year = {2018},
    pages = {856--865},
}

@inproceedings{rajpurkar-etal-2016-squad,
    title = "{SQ}u{AD}: 100,000+ Questions for Machine Comprehension of Text",
    author = {Rajpurkar, Pranav  and
      Zhang, Jian  and
      Lopyrev, Konstantin  and
      Liang, Percy},
    booktitle = {EMNLP},
    year = {2016},
    pages = {2383--2392}
}

@inproceedings{morris-etal-2020-textattack,
    title = "{T}ext{A}ttack: A Framework for Adversarial Attacks, Data Augmentation, and Adversarial Training in {NLP}",
    author = "Morris, John  and
      Lifland, Eli  and
      Yoo, Jin Yong  and
      Grigsby, Jake  and
      Jin, Di  and
      Qi, Yanjun",
    booktitle = "EMNLP: System Demonstrations",
    year = "2020",
    pages = "119--126",
}

@inproceedings{alzantot-etal-2018-generating,
    title = "Generating Natural Language Adversarial Examples",
    author = "Alzantot, Moustafa  and
      Sharma, Yash  and
      Elgohary, Ahmed  and
      Ho, Bo-Jhang  and
      Srivastava, Mani  and
      Chang, Kai-Wei",
    booktitle = "EMNLP",
    pages = "2890--2896",
}

@inproceedings{DBLP:conf/icde/MalekpourRLM26,
  author       = {Mohammadhossein Malekpour and
                  Mohamed Riahi and
                  Maxime Lamothe and
                  Amine Mhedhbi},
  title        = {SQLMorph: Query Mutation and Fine-Grained Metrics for Text-to-SQL
                  Evaluation},
  booktitle    = {ICDE},
  pages        = {2628--2640},
  year         = {2026},
}

@inproceedings{NEURIPS2025_d077bc9e,
 author = {Liang, Buyun and Peng, Liangzu and Luo, Jinqi and Thaker, Darshan and Chan, Kwan Ho Ryan and Vidal, Rene},
 booktitle = {NeurIps},
 doi = {10.52202/085713-4753},
 pages = {142059--142099},
 title = {SECA: Semantically Equivalent and Coherent Attacks for Eliciting LLM Hallucinations},
 year = {2025}
}

@inproceedings{NEURIPS2024_f5198bc2,
 author = {Zhang, Zhehao and Chen, Jiaao and Yang, Diyi},
 booktitle = {NeurIps},
 doi = {10.52202/079017-4317},
 pages = {135904--135942},
 title = {DARG: Dynamic Evaluation of Large Language Models via Adaptive Reasoning Graph},
 year = {2024}
}

@InProceedings{pmlr-v267-paulus25a,
  title = 	 {{A}dv{P}rompter: Fast Adaptive Adversarial Prompting for {LLM}s},
  author =       {Paulus, Anselm and Zharmagambetov, Arman and Guo, Chuan and Amos, Brandon and Tian, Yuandong},
  booktitle = 	 {ICML},
  pages = 	 {48439--48469},
  year = 	 {2025},
}

@inproceedings{oza-etal-2026-llms,
    title = "{LLM}s are Brittle to Simple Code Transformations: Introducing {CETB}ench {--} A Benchmark for Code-Equivalence Checking",
    author = "Oza, Neeva  and
      Govil, Ishaan  and
      Gupta, Parul  and
      Khandelwal, Dinesh  and
      Garg, Dinesh  and
      Singla, Parag",
    booktitle = "Findings of the Association for Computational Linguistics: ACL",
    year = "2026",
    pages = "41653--41685",
}

@inproceedings{kuhn2023semantic,
  title = {Semantic Uncertainty: Linguistic Invariances for Uncertainty Estimation in Natural Language Generation},
  author = {Kuhn, Lorenz and Gal, Yarin and Farquhar, Sebastian},
  booktitle = {International Conference on Learning Representations},
  year = {2023},
  url = {https://openreview.net/forum?id=VD-AYtP0dve}
}

@article{farquhar2024detecting,
  title = {Detecting Hallucinations in Large Language Models Using Semantic Entropy},
  author = {Farquhar, Sebastian and Kossen, Jannik and Kuhn, Lorenz and Gal, Yarin},
  journal = {Nature},
  volume = {630},
  number = {8017},
  pages = {625--630},
  year = {2024},
  doi = {10.1038/s41586-024-07421-0}
}

@inproceedings{chen-etal-2021-finqa,
  title = {{F}in{QA}: A Dataset of Numerical Reasoning over Financial Data},
  author = {Chen, Zhiyu and Chen, Wenhu and Smiley, Charese and Shah, Sameena and Borova, Iana and Langdon, Dylan and Moussa, Reema and Beane, Matt and Huang, Ting-Hao and Routledge, Bryan and Wang, William Yang},
  booktitle = {Proceedings of the 2021 Conference on Empirical Methods in Natural Language Processing},
  year = {2021},
  pages = {3697--3711},
  publisher = {Association for Computational Linguistics},
  doi = {10.18653/v1/2021.emnlp-main.300}
}

@inproceedings{jin-etal-2019-pubmedqa,
  title = {{P}ub{M}ed{QA}: A Dataset for Biomedical Research Question Answering},
  author = {Jin, Qiao and Dhingra, Bhuwan and Liu, Zhengping and Cohen, William and Lu, Xinghua},
  booktitle = {Proceedings of the 2019 Conference on Empirical Methods in Natural Language Processing and the 9th International Joint Conference on Natural Language Processing},
  year = {2019},
  pages = {2567--2577},
  publisher = {Association for Computational Linguistics},
  doi = {10.18653/v1/D19-1259}
}

@inproceedings{koreeda-manning-2021-contractnli,
  title = {{C}ontract{NLI}: A Dataset for Document-level Natural Language Inference for Contracts},
  author = {Koreeda, Yuta and Manning, Christopher},
  booktitle = {Findings of the Association for Computational Linguistics: EMNLP 2021},
  year = {2021},
  pages = {1907--1919},
  publisher = {Association for Computational Linguistics},
  doi = {10.18653/v1/2021.findings-emnlp.164}
}

@inproceedings{guha2023legalbench,
  title = {{L}egal{B}ench: A Collaboratively Built Benchmark for Measuring Legal Reasoning in Large Language Models},
  author = {Guha, Neel and Nyarko, Julian and Ho, Daniel E. and R{\'e}, Christopher and Chilton, Adam and Narayana, Aditya and Chohlas-Wood, Alex and Peters, Austin and Waldon, Brandon and Rockmore, Daniel N. and others},
  booktitle = {Advances in Neural Information Processing Systems 36, Datasets and Benchmarks Track},
  year = {2023}
}

@article{phogat2023zeroshot,
  author = {Phogat, Karmvir Singh and Harsha, Chetan and Dasaratha, Sridhar and Ramakrishna, Shashishekar and Puranam, Sai Akhil},
  title = {Zero-Shot Question Answering over Financial Documents using Large Language Models},
  journal = {arXiv preprint arXiv:2311.14722},
  year = {2023},
}

@inproceedings{schuster2022stretching,
  author = {Schuster, Tal and Chen, Sihao and Buthpitiya, Senaka and Fabrikant, Alex and Metzler, Donald},
  title = {Stretching Sentence-pair {NLI} Models to Reason over Long Documents and Clusters},
  booktitle = {Findings of the Association for Computational Linguistics: EMNLP 2022},
  year = {2022},
}

@article{singhal2023clinical,
  author = {Singhal, Karan and Azizi, Shekoofeh and Tu, Tao and Mahdavi, S. Sara and Wei, Jason and Chung, Hyung Won and Scales, Nathan and Tanwani, Ajay and Cole-Lewis, Heather and Pfohl, Stephen and others},
  title = {Large language models encode clinical knowledge},
  journal = {Nature},
  volume = {620},
  pages = {172--180},
  year = {2023},
}

@article{nori2023medprompt,
  author = {Nori, Harsha and Lee, Yin Tat and Zhang, Sheng and Carignan, Dean and Edgar, Richard and Fusi, Nicolo and King, Nicholas and Larson, Jonathan and Li, Yuanzhi and Liu, Weishung and others},
  title = {Can Generalist Foundation Models Outcompete Special-Purpose Tuning? Case Study in Medicine},
  journal = {arXiv preprint arXiv:2311.16452},
  year = {2023},
}

@book{efron1993bootstrap,
  title={An Introduction to the Bootstrap},
  author={Efron, Bradley and Tibshirani, Robert J.},
  year={1993},
  publisher={Chapman \& Hall/CRC}
}

@misc{pngkeyMetapod,
  author       = {{PNGKey}},
  title        = {011metapod Dream -- Pokemon Metapod},
  howpublished = {\url{https://www.pngkey.com/maxpic/u2q8u2o0e6t4t4o0/}},
  note         = {Accessed September 18, 2026}
}

\newpage


\appendix

\section{HARDEN artifacts and check validation}
\label{app:degradation-artifacts}

This appendix records concrete artifacts from our experiments.  We include
them to make the information boundaries, calibration decisions, and check
behavior auditable rather than describing them only at the level of prompts.

\subsection{Example target-blind realism rubric}
\label{app:rubric-example}

Table~\ref{tab:finqa-rubric} summarizes an actual rubric created by HARDEN's realism
check for FinQA.  The domain-specific rubric was first researched, finding TAT-QA and FinanceBench as external benchmark
families and SEC EDGAR filings as a primary real-world source family, after which it calibrated against real FinQA benchmark cases.  The
table preserves the rubric's operational thresholds and high-score anchors
while shortening the descriptions for presentation.

\begin{table*}[t]
\caption{The target-blind ``Financial-report numerical-reasoning input
realism rubric'' used in our FinQA experiment.  Scores use a 1--5 scale.}
\label{tab:finqa-rubric}
\centering
\small
\begin{tabular}{p{0.07\textwidth}p{0.22\textwidth}p{0.60\textwidth}}
\toprule
ID & Dimension & Operational high-score anchor \\
\midrule
D1 & Tabular numeric density and structure &
A genuine financial table, typically with at least eight labeled line items,
at least two value columns, and at least 25\% numeric cells. \\
D2 & Financial line-item and disclosure vocabulary &
At least six distinct financial terms or line items, such as revenue, cost of
sales, operating income, total assets, cash flow, or fiscal year. \\
D3 & Authentic accounting numeric formatting &
At least three authentic conventions, including thousands separators,
parenthesized negatives, currency or scale markers, period headers,
percentages, per-share values, or footnote markers. \\
D4 & Grounded numerical-reasoning intent &
The question names a metric or period present in the report and requests a
computable value, change, difference, ratio, percentage, or aggregate. \\
\bottomrule
\end{tabular}
\end{table*}

A case passes only when its mean score is at least 4 out of 5, every dimension is at
least 4 out of 5, and no blocking red flag is detected.  The two blocking flags are
(R1) an absent or degenerate financial table and (R2) numeric cells with no
authentic accounting convention.  The rubric also records two non-blocking
signals that lower the relevant dimension scores: generic placeholder
vocabulary (R3) and a question ungrounded in the supplied report (R4).

Applied diagnostically to the original and synthetic (20 each generated by GPT-5.6 Sol and GPT-3.5 Turbo respectively) FinQA cases used in the
check-validation study, the frozen rubric's per-source acceptance rates are
those reported in Table~\ref{tab:realism-source-rates}
(Appendix~\ref{app:check-validation}); 19 of the 20 rejected GPT-3.5 Turbo
cases triggered the degenerate-table flag. 

\subsection{Example complexity-axis catalog}
\label{app:axis-example}

Table~\ref{tab:finqa-axes} shows the complete set of ten axes produced by the
domain-research stage for FinQA.  Each axis specifies a family of realistic,
target-preserving mutations rather than one fixed transformation.

\begin{table*}[t]
\caption{Generated FinQA complexity-axis catalog, with abbreviated examples.}
\label{tab:finqa-axes}
\centering
\small
\begin{tabular}{p{0.29\textwidth}p{0.62\textwidth}}
\toprule
Axis & Example mutation \\
\midrule
Table-label compression &
Replace verbose row and column descriptions with authentic filer shorthand. \\
Table-lookup ambiguity &
Add plausible sibling rows, subtotals, or comparative-period columns. \\
Numeric-format heterogeneity &
Mix currency symbols, separators, parenthetical negatives, and dash zeros. \\
Financial-terminology variation &
Use standard financial synonyms across the question and source. \\
Narrative-distractor density &
Add non-decisive figures and footnote detail from related disclosures. \\
Cross-reference indirection &
Route a fact through a resolvable footnote or remove a duplicate restatement. \\
Table-structure reformatting &
Reorder rows or restructure period headers while preserving values. \\
Temporal-period ambiguity &
Mix fiscal/calendar references and authentic period-bucket labels. \\
Filing-capture artifacts &
Add sparse PDF-to-text artifacts only to non-critical spans. \\
Domain-specific realism &
Coherently combine several filing-authentic complexity mechanisms. \\
\bottomrule
\end{tabular}
\end{table*}

\subsection{Example original and \textsc{HARDEN}-generated case}
\label{app:harden-example}

Table~\ref{tab:harden-example} excerpts a ContractNLI case hardened by \textsc{HARDEN} during
the completed full run under the legal-boilerplate and recital-inflation
axis.  The evaluation target is unchanged: the hypothesis ``Receiving Party
shall destroy or return some Confidential Information upon the termination of
Agreement.''\ retains its gold label \texttt{NotMentioned}, and the operative
clauses of the agreement are preserved verbatim.  The added material is
ceremonial and non-determinative, yet the mean Qwen~3.5 397B score over ten
trials falls from 1.0 to 0.3.

\begin{table*}[t]
\caption{Excerpt of a ContractNLI case before and after evolution.  The
mutation prepends authentic recitals and pads the execution formalities of a
real mutual NDA; ellipses mark elided unchanged text.}
\label{tab:harden-example}
\centering
\scriptsize
\begin{tabular}{p{0.46\textwidth}p{0.46\textwidth}}
\toprule
Original & \textsc{HARDEN} \\
\midrule
MUTUAL NON-DISCLOSURE AGREEMENT\newline
Each undersigned party (the ``Receiving Party'') understands
that\ldots &
MUTUAL NON-DISCLOSURE AGREEMENT\newline
\textbf{WITNESSETH:}\newline
\textbf{WHEREAS}, CombiChem, Inc., a Delaware corporation, and DuPont
Pharmaceuticals Company, a Delaware general partnership (each a ``party''
and collectively the ``parties''), are engaged in discussions concerning a
possible business transaction of mutual interest; and\newline
\textbf{WHEREAS}, in the course of such discussions each party may find it
desirable to disclose to the other certain of its proprietary and
confidential information\ldots; and\newline
\textbf{WHEREAS}, the parties desire to establish the terms and conditions
governing the disclosure, protection and use of such information;\newline
\textbf{NOW, THEREFORE}, in consideration of the mutual covenants and
promises contained herein\ldots, the parties agree as set forth
below.\newline
Each undersigned party (the ``Receiving Party'') understands
that\ldots \\
\addlinespace
IN WITNESS WHEREOF, the parties have executed this Agreement as of the day
and year set forth below.\newline
Date: 3-10-99 \ldots &
IN WITNESS WHEREOF, the parties have \textbf{caused this Agreement to be
executed by their respective duly authorized representatives} as of the day
and year set forth below.\newline
Date: 3-10-99 \ldots\newline
\textbf{Address for Notices:}\newline
\textbf{CombiChem, Inc., Attn: General Counsel, San Diego,
California}\newline
\textbf{DuPont Pharmaceuticals Company, Attn: Legal Department, Wilmington,
Delaware} \\
\bottomrule
\end{tabular}
\end{table*}

\subsection{Search and task model details}
\label{app:runtime-details}

Rubric research receives the use-case description, schemas, execution plan, and
system prompt, but no target cases, labels, mutations, evaluator results, or
task model scores.  It draws on at least two independent external benchmark
families and one primary real-world source family, calibrates 3--5 realism dimensions on
held-out examples, and freezes the rubric by SHA-256
hash.  Complexity axis discovery is separate and may inspect target inputs, immutable
outputs, and the evaluator to identify label-preserving difficulty mechanisms.
During evolution, HARDEN records mutation axes applications that pass
both LM-based checks and lower fitness; it rewards prior improvement and
penalizes repeated non-improvement.

Each benchmark is evaluated with the Qwen~3.5 35B, 122B, and 397B task
models served through OpenRouter with reasoning disabled. Ten sampled
completions use pinned model-card profiles in \ref{subsec:search-model-toolconfig}; these samples are used for primary performance and uncertainty.  A temperature-zero completion and its token log-probabilities are retained for post-hoc sequence NLL and mean-token NLL. OpenRouter returns output-token
log-probabilities for all generations, which the post-hoc likelihood-based uncertainty measures require.

In a preliminary evaluation using 15 official cases per benchmark and two independent calls per setting, enabling reasoning produced no consistent accuracy benefit. On FinQA with the 35B model, accuracy decreased by 3.3\%, from 76.7\% to 73.3\%. On PubMedQA with the 122B model, accuracy increased by 3.3\%, from 46.7\% to 50.0\%. These mixed effects came with substantially greater token usage: approximately 3.7$\times$ on FinQA (864 to 3.2k tokens) and 8.3$\times$ on PubMedQA (228 to 1.9k tokens), as well as increased latency. Further, the output token log-probabilities referenced to calculate uncertainty as a secondary measure of task difficulty are incompatible with reasoning models. This is due to the fact that the output token lob-probabilities would be conditioned on input tokens as well as reasoning tokens, which differ across samples; this confounds the task uncertainty we aim to measure.

\paragraph{Reproducibility materials.}
Appendix~\ref{app:reproducibility-prompts} discloses the fixed method prompts,
dynamic-field schemas, search configuration, model settings, and tool
boundaries used by the experiments.  No code, data, or exact run outputs are
released.  The disclosure is intended to support an independent implementation
and reproduction of the experimental procedure.

Our post-hoc likelihood-weighted semantic entropy follows
Kuhn et al.~\citep{kuhn2023semantic}: samples are grouped into exact task-level semantic
classes, weighted by length-normalized sequence likelihood, and aggregated
before computing entropy.  Online selection instead uses a normalized
adaptation of the discrete semantic entropy introduced by
Farquhar et al.~\citep{farquhar2024detecting}, estimating class probabilities from sample
frequencies.  For PubMedQA and ContractNLI, the exact semantic classes are the
three answer labels,
$u_i(x_i)=-\sum_z\hat p_z\log\hat p_z/\log 3$.  For FinQA, $z$ is the distinct
executed numerical result and the normalization is $\log T$, where $T=10$.
Post-hoc analysis
also reports sequence NLL, mean-token NLL, frequency entropy, and
likelihood-weighted entropy; these complementary measures characterize
confidence without changing the primary task-score objective.

\subsection{Reproducibility prompt and configuration disclosure}
\label{app:reproducibility-prompts}

This appendix discloses the method-critical templates and dynamic-field
schemas used in the experiments.  Line wrapping and surrounding configuration
syntax are normalized for \LaTeX{}.  Internal terminology is rendered as
``task model'' for consistency.
The verbatim production templates retain the implementation term ``gate'';
it denotes the correctness or realism check described in the paper, not a
separate mechanism.

\subsubsection{Search, model, and tool configuration}
\label{subsec:search-model-toolconfig}

The paper-facing search configuration samples 200 cases without replacement
with seed 42, preserves source order after sampling, and partitions the cases
into ten deterministic worker runs.  The same mutation axis catalog is shared across worker runs.

Rubric research, axis research, and mutation identify Claude Opus~4.8
(\texttt{anthropic/claude-opus-4-8}) with adaptive thinking and high effort.
No temperature or output-token override is applied to those agent-driven
calls, so they use the agent runtime defaults.  Correctness (both extraction and
audit) and realism (audit) calls use the same model through direct OpenRouter checks
with temperature 0, at most 16{,}384 output tokens, JSON-object responses, and
high reasoning effort.  They fail closed unless the response metadata confirms that route.  The three-member ensembles use majority voting, which
requires two of three passes.

All task-model profiles request one greedy completion at temperature 0 and
ten sampled completions.  Reasoning is disabled, repetition penalty is 1.0,
maximum output tokens are 16{,}384, and selected-token log probabilities
request the top five alternatives.  OpenRouter parameter enforcement is
required.  The exact sampled profiles are:
\begin{itemize}
\item \texttt{qwen/qwen3.5-35b-a3b}: Parasail, temperature 1.0,
  top-$p$ 0.95, top-$k$ 20, and presence penalty 1.5.
\item \texttt{qwen/qwen3.5-122b-a10b}: Novita, temperature 1.0,
  top-$p$ 1.0, top-$k$ 40, and presence penalty 2.0.
\item \texttt{qwen/qwen3.5-397b-a17b}: Parasail, temperature 0.7,
  top-$p$ 0.8, top-$k$ 20, and presence penalty 1.5.
\end{itemize}

Rubric research may read, search, fetch, and use shell commands, while the
named target benchmark family and its mirrors and derivatives are forbidden.
Axis research may use the web.  Before inspecting the benchmark cases, the axis researcher reads two fixed support files: a research specification requiring evidence-backed, real-world difficulty that preserves labels and avoids rubric-targeting, and a catalog of generic transformation ideas used only as seeds for domain-specific axis discovery. Their paths are included in the single axis-research request; they do not create additional model calls.  Mutation
generation has no web, read, search, or shell access and works only from the
constructed context.  Direct checks and task-model calls use no tools.
Both single-pass baselines use Claude Opus 4.8 as the mutation model with adaptive thinking, high reasoning effort, default temperature, one user turn, and a maximum of 16,384 output tokens.

\paragraph{Deterministic selection and artifacts.}
Parents and check-approved children are sorted by ascending task-model score.
Exact score ties prefer higher normalized sampled-outcome uncertainty and then
a new child.  The five best survive.  A score of zero is treated as collapsed
and rejected.  Final promotion requires a strict score improvement over the
immutable baseline.  Before scoring, deterministic checks require a valid
mutation index and manifest, explicit passes from both correction and realism checks, edits confined
to declared mutable inputs and unchanged immutable files.  Finalization reconstructs each selected case from its immutable
snapshot, copies only declared mutable artifacts, parses selected JSON files,
and runs the structural completion check.  These are deterministic code
contracts, not model prompts.

\subsubsection{Rubric research templates}

Prompt~A.1 is the system instruction for the target-blind rubric researcher.
Prompt~A.2 is the user request sent to that researcher once before case-level
evolution begins.  They are the two messages in one rubric-research call.
\begin{verbprompt}{Rubric researcher system instruction}
You are a domain-realism research agent. Build an evidence-backed realism rubric that can judge whether benchmark inputs resemble authentic records from the relevant real-world domain.

The parent workflow supplies a bundle root and use case. You MUST read `grounding/domain_realism/research_constraints.json`. Every degradation rubric run is target-blind: infer the domain and task family ONLY from metadata, schemas, prompts, and the generic use case. The research bundle intentionally contains no target cases. Do not search for, retrieve, read, cite, normalize, or calibrate on examples from any family named in `excluded_benchmark_families`, including mirrors and derivative datasets.

Never use expected outputs, labels, gold answers, evaluator results, mutation artifacts, or task-model scores as realism evidence.

This prompt is intentionally domain-agnostic. You must discover the relevant reference sources yourself. Do not assume that the target benchmark's formatting conventions define domain realism.

Research requirements:
1. Discover and retrieve samples from at least two EXTERNAL, INDEPENDENT benchmark families relevant to the inferred domain and task. The target benchmark does not count toward this minimum. Datasets derived from the same underlying records count as one family.
2. Retrieve samples from at least one PRIMARY real-world source for the domain. A benchmark, blog, paper description, or synthetic corpus is not primary real-world data.
3. Verify source relevance, lineage, canonical URL, revision/version, and license or usage provenance. Reject weakly related sources.
4. Save immutable snapshots of every sample used. Normalize INPUT artifacts into a common readable representation while retaining the raw snapshots.
5. Exclude labels, expected answers, gold programs, evaluator rubrics, and model outputs from normalized research samples.
6. Derive common domain properties across sources. Keep source-specific serialization quirks separate. A blocking criterion needs evidence from at least two independent source families, or one benchmark family plus primary real-world data.
7. Build 3-5 authenticity dimensions with a 1-5 scale, concrete high/low anchors, a met threshold, synthetic red flags, and an explicit case-level pass rule. Do not score task correctness.
8. Calibrate the rubric on deterministic, disjoint probe and held-out samples from every source family. Real held-out records should meet the rubric; do not weaken it merely to force calibration.
9. In a target-blind run, every registry source must set `is_target_family: false`; excluded target examples must not appear anywhere in `sources/`, the registry, reference profile, calibration, or report.

Write everything below `${BUNDLE_ROOT}/grounding/domain_realism/`:
- `source_registry.json`
- `rubric.json`
- `reference_profile.json`
- `research_report.md`
- `sources/` containing raw and normalized sample snapshots

`source_registry.json` must be:
{
  "inferred_domain": "concise domain",
  "inferred_task_family": "concise task family",
  "target_benchmark_family": "family name or unknown",
  "sources": [
    {
      "id": "stable_source_id",
      "name": "source name",
      "family": "independent lineage family",
      "kind": "benchmark or primary_real_world",
      "is_target_family": false,
      "canonical_url": "https://...",
      "revision": "commit, version, date, or immutable identifier",
      "license_or_provenance": "license or access provenance",
      "retrieved_at": "ISO-8601 timestamp",
      "relevance": "why this source matches the inferred domain and task",
      "lineage": "underlying record lineage and known derivatives",
      "raw_sample_paths": ["grounding/domain_realism/sources/..."],
      "normalized_sample_path": "grounding/domain_realism/sources/...",
      "sample_count": 10,
      "sha256": "sha256 of the normalized sample file"
    }
  ]
}

`rubric.json` must be:
{
  "rubric_name": "domain-specific name",
  "domain_summary": "holistic synthesis shared across sources",
  "scale_min": 1,
  "scale_max": 5,
  "met_threshold": 4,
  "dimensions": [
    {
      "id": "D1",
      "name": "dimension name",
      "description": "what authenticity property is judged",
      "score_high": "concrete anchor for 5",
      "score_low": "concrete anchor for 1",
      "evidence_source_ids": ["source_a", "source_b"]
    }
  ],
  "synthetic_red_flags": [
    {
      "id": "R1",
      "description": "blocking or strong synthetic signal",
      "blocking": true,
      "evidence_source_ids": ["source_a", "source_b"]
    }
  ],
  "case_pass_rule": {
    "minimum_mean_score": 4.0,
    "all_dimensions_must_meet_threshold": true,
    "blocking_red_flags_fail": true
  },
  "overall_guidance": "how to apply the common rubric without overfitting one source",
  "calibration": {
    "probe_paths": ["grounding/domain_realism/sources/..."],
    "heldout_paths": ["grounding/domain_realism/sources/..."],
    "heldout_results_by_source": {
      "source_id": {"n": 5, "mean_score": 4.2, "met_rate": 0.8}
    },
    "target_met": true
  }
}

`reference_profile.json` must identify common patterns, source-specific
patterns, deterministic measurements, and the normalized examples the later
independent auditor may use. Every claim must cite source IDs.

Use absolute paths for file operations. Finish only after all required files
exist and parse. Reply with a concise source/rubric summary; the workflow reads
the files directly.
\end{verbprompt}

The uppercase names in the request are dynamic fields.
\begin{verbprompt}{Rubric research user request}
BUNDLE_ROOT: ${bundle_dir}

USE_CASE: ${use_case}

Deep-research a domain-realism rubric using the generic requirements in your
agent instructions. Infer the domain and task from this bundle. Discover all
reference sources yourself; no source names or domain-specific rubric
dimensions are being supplied by the workflow. Use absolute paths rooted at
BUNDLE_ROOT and write every required artifact under
BUNDLE_ROOT/grounding/domain_realism/.
\end{verbprompt}

The dynamic fields are \texttt{bundle\_dir} and \texttt{use\_case}.
Deterministic validation runs before the rubric is frozen.  If validation
fails, its errors are returned to the researcher for correction before strict
validation is repeated; that conditional repair instruction is not a separate
experimental prompt.

\subsubsection{Axis research templates}

The following system instruction and user request form the axis-research call
that produces one catalog shared by all worker runs.
\begin{verbprompt}{Axis researcher system instruction}
You are the degradation researcher for a completed Button step01 benchmark bundle.

Your job is to research how to add realistic noise and obscurity to this use
case's inputs, then design degradation axes. Every case in the bundle is in
scope for degradation; do not classify, prioritize, or exclude cases as
non-degradable. When invoked by the shared-axis research workflow, the bundle
is the full corpus and your axes become the immutable pool for all execution
lanes. Do not modify cases, executor.py, evaluator.py, prompts/, resources/, or
gold outputs.

Definitions for current Button bundles:
- `input_data`: `cases/case_N/input.json` plus files listed in
  `cases/case_N/metadata.json` under `input`. These are the only degradation
  targets.
- `gold_output`: `cases/case_N/output.json` plus files listed in metadata
  under `expected_output`. These are immutable.
- `shared_resources`: `resources/` and `prompts/`. These are immutable.
- Degradable text artifacts: `.json`, `.csv`, `.txt`, `.html`, `.md`, `.xml`, `.yaml`, `.yml`.
- Immutable artifacts: images, audio, video, PDFs without an editable text sidecar, archives, binaries.

Hard-but-real filter for every axis:
- Real-world mechanism: what ordinary process produces this complexity?
- Plausibility: would this appear in raw real inputs for this domain?
- Rubric independence: would it still be hard without knowing the evaluator rubric?
- Organic over manufactured: prefer messy workflows, institutional variation,
  corrections, ambiguity, density, and implicit context; reject gotchas,
  planted contradictions, and removed facts.

Workflow:
1. Before reading the bundle, read both required paths supplied by the parent workflow. This is mandatory:
   - `RESEARCH_DOSSIER_PROMPT_PATH`: the hard-case research dossier method.
     Apply its real-world-mechanism, rubric-independence, label-validity,
     evidence-class, and anti-input-hacking requirements to this research.
   - `GENERIC_AXIS_SEED_PATH`: generic transformation ideas. Treat them only
     as a seed vocabulary; do not copy their names, definitions, or prompt
     guidance as final axes. Derive domain-specific axes from the bundle and
     researched evidence, and reject any seed that lacks a plausible domain
     mechanism.
   - If either path is unset or unreadable, stop and report the missing path
     rather than proceeding without it. Cite both paths in
     `degradation/degradation_research_report.md`, including which dossier
     constraints and seed concepts informed or were rejected from the final
     axes.
2. Read `bundle_manifest.json`, `grounding/authoring_digest.json`,
`grounding/realism_brief.json`, `grounding/realism_criteria.json`,
`executor.py`, `evaluator.py`, `prompts/*.j2`, `benchmark_results.json`, and
all `cases/case_*/metadata.json`, `input.json`, and `output.json` files. Treat
every input/gold pair in this research bundle as the evidence base for
selecting axes: prioritize mechanisms that fit one or more examples while
preserving every immutable gold output, and reject axes that are only generic
domain ideas with no plausible application to the corpus.
3. Read any case input sidecar files listed by metadata that are text artifacts.
4. For every case, record the input artifacts to mutate, relative to the case
directory, such as `input.json` or `invoice.csv`. Include every case in
`degradable_cases`; do not create researcher-level skipped cases.
5. Research domain-specific difficulty patterns. Use web search if available;
persist useful fetched evidence under `degradation/retrieved_*.md`.
6. Write all of these files:
   - `degradation/degradation_axes.json`: every evidence-backed reusable axis
     discovered from this corpus; do not impose an arbitrary axis-count cap.
     Exactly one axis must be named `domain_specific_realism`. Each axis must
     include `axis_name`, `definition`, `goal`, `distinctive_focus`,
     `target_artifact_types`, at least 3 `subtypes`, and at least 5
     `prompt_guidance` strings. Every axis must explicitly say it operates only
     on case input artifacts and must preserve gold output correctness. Add
     optional `priority` (`high`, `normal`, or `low`) based on expected
     effectiveness for the corpus. Select axes that add plausible noise,
     ambiguity, obscurity, or retrieval burden to concrete examples; do not
     produce generic axes disconnected from the inputs. For FinCoT-like
     financial tables, include high-priority table-label compression and
     table/lookup ambiguity axes when evidence supports them; de-prioritize
     broad prose or OCR-only changes unless they preserve a meaningful
     table-interpretation challenge.
   - `degradation/degradation_research_report.md`
   - `degradation/degradation_research_context.json`
   - `degradation/manifest.json`
   - `degradation/degradable_cases.json` - this exact JSON shape:
{
  "degradable_cases": [
    {
      "case_id": "case_1",
      "baseline_score": 0.92,
      "degradable_files": ["input.json", "invoice.csv"]
    }
  ],
  "skipped_cases": [],
  "summary": "Short prose summary of what you found."
}

`baseline_score` must be normalized to `[0, 1]`.

After writing every file, reply with a brief plain-text summary of what you
produced. The workflow reads `degradation/degradable_cases.json` from disk.
\end{verbprompt}

The following request is sent once to produce the axis catalog shared by all
worker runs.
\begin{verbprompt}{Shared-axis research user request}
Bundle root: ${bundle_dir}

Use case: ${use_case}

This is the single shared-axis research stage for a partitioned degradation
run. The bundle contains the complete corpus that all later lanes will process.
Read both required context files before examining the bundle:

RESEARCH_DOSSIER_PROMPT_PATH: ${research_dossier_prompt_path}
GENERIC_AXIS_SEED_PATH: ${generic_axis_seed_path}

Curate one reusable global degradation-axis catalog from the full corpus.
Return every evidence-backed axis that is plausible for one or more cases; do
not impose an arbitrary axis-count cap. Write all standard research artifacts
under ${bundle_dir}/degradation/. This stage researches only: do not invoke
evolutionary initialization, mutation, evaluation, or finalization.
\end{verbprompt}

The dynamic fields are \texttt{bundle\_dir}, \texttt{use\_case},
\texttt{research\_dossier\_prompt\_path}, and
\texttt{generic\_axis\_seed\_path}.  The dossier and seed are context files,
not prompts attributed to the axis researcher.

\subsubsection{Mutation template and context schema}
\label{app:mutation-template}

\paragraph{Axis ranking.}
For an axis $a$ with $A_a$ prior applications, $G_a$ constraint-passing
applications, and $I_a$ fitness-improving applications, the base priority is
$4I_a/A_a + G_a/A_a$, or $1.5$ when $A_a=0$.
The implementation adds $3$ for
axes assigned high priority during research. It subtracts $4$ after at least two applications without improvement, $2$ when
the constraint-passing rate is below $0.3$ after at least two applications, and $1$
when the axis already appears in the retained lineage. It may also add benchmark-specific bonuses: in the FinQA runs, it additionally adds $5$ for table-label compression axes and $4$ for table-lookup. Axes are sorted by
this score.  The mutation agent receives the ranking and underlying counts,
then selects one axis per offspring while preferring distinct, previously
unused axes.

The following system instruction and user request form each mutation-generation
call.
\begin{verbprompt}{Mutation generator system instruction}
You are the mutation generator in an independently gated evolutionary
degradation pipeline.

Generate and write candidate degraded versions of one benchmark case. You do
not judge correctness, realism, or fitness. Separate independent gates own all
acceptance decisions.

The workflow prompt supplies `MUTATION_CONTEXT_JSON` with the current case,
candidate directories, writable files, gold-output summary, degradation axes,
evolutionary feedback, and a manifest path. Use only that context. Write only
inside the supplied candidate directories and manifest path.

Generation requirements:
1. Identify every load-bearing fact, value, relationship, and constraint
   needed for the immutable expected output.
2. Read `candidates_per_generation` and produce exactly that many candidates.
3. Use exactly one supplied degradation axis per candidate. Prefer distinct,
   high-performing axes not already in the mutation history.
4. Apply plausible real-world complexity, not planted traps, contradictions,
   or deletion of necessary information.
5. Write complete replacement content to each assigned `case_dir`, touching
   only its `writable_files`.
6. Do not run the executor/evaluator and do not predict whether any gate will pass.

Write the manifest to `manifest_path`:
{
  "mutations": [
    {
      "individual_index": 0,
      "axes": ["axis_name"],
      "subtypes": ["subtype_name"],
      "description": "What changed and its plausible real-world origin.",
      "written_files": ["input.json"]
    }
  ]
}

Use integer indexes from zero through `candidates_per_generation - 1`, exactly
once each. Never modify expected outputs, metadata, prompts, resources,
executor/evaluator code, or canonical case files.

After writing all candidates and the manifest, reply only with a concise count and the axes used.
\end{verbprompt}

Each mutation call supplies
\texttt{MUTATION\_CONTEXT\_JSON: \{context\}} followed by:
\begin{verbprompt}{Mutation generation user request}
Use only this context. Generate exactly `candidates_per_generation` mutations,
write complete replacements to the assigned candidate directories, and write
the manifest to `manifest_path`. Do not judge correctness, realism, or fitness.
\end{verbprompt}

The deterministic mutation context supplies the following exhaustive
top-level fields:
\texttt{case\_id}, \texttt{generation}, \texttt{baseline\_score},
\texttt{candidates\_per\_generation}, \texttt{population\_size},
\texttt{max\_generations}, \texttt{degradable\_files},
\texttt{source\_case\_dir}, \texttt{candidates},
\texttt{manifest\_path}, \texttt{case\_state},
\texttt{candidate\_axes}, \texttt{axis\_feedback},
\texttt{realism\_criteria\_summary}, \texttt{metadata},
\texttt{gold\_output\_summary}, and \texttt{files}.
Each \texttt{candidates} entry contains \texttt{individual\_index},
\texttt{case\_dir}, and \texttt{writable\_files}.  When present,
\texttt{case\_state} contains \texttt{current\_generation},
\texttt{best\_score}, \texttt{best\_individual\_path}, and
\texttt{population}; each population entry contains \texttt{score},
\texttt{path}, \texttt{axes}, \texttt{mutation\_chain}, and
\texttt{description}.  Each candidate axis contains \texttt{axis\_name},
\texttt{priority}, \texttt{definition}, \texttt{goal},
\texttt{distinctive\_focus}, \texttt{subtypes}, and
\texttt{prompt\_guidance}.  Each axis-feedback entry contains
\texttt{applied}, \texttt{survived\_gates}, and
\texttt{improved\_fitness}.

\subsubsection{Correctness extraction and audit prompts}

The correctness check makes one extraction call for the parent case and one
audit call for each candidate.  The boxes below show the complete user
requests after fixed instructions and dynamic input fields are combined.
\begin{verbprompt}{Correctness information extraction request}
Respond with a single valid JSON object only. Do not include Markdown, prose, or analysis.

<TASK>
Identify the minimal set of information necessary to produce the expected output.
</TASK>

<BASELINE_EXECUTION_CONTEXT>
The baseline execution task is:
{{ baseline_execution_prompt }}
</BASELINE_EXECUTION_CONTEXT>

<INSTRUCTIONS>
Given an input, supplementary data, and expected output, identify the MINIMAL set of information
from the INPUT that is NECESSARY to reason about and produce the expected output.

IMPORTANT: Extract information ONLY from the INPUT, not from supplementary or output.
- The supplementary data provides context (e.g., policies, guidelines) for decision-making
- The expected output shows what decision was made
- Your job: identify which pieces of INPUT information were critical for reaching that output

CRITICAL REQUIREMENT - Output Evidence:
Before marking any information as "necessary", you MUST:
1. Find the EXACT text in the expected output that depends on this information
2. Quote the specific words/phrases from the output
3. Explain the direct causal link: input information -> reasoning -> output text

DO NOT use vague terms like "implies", "suggests", "references", "indicates", "conveys".
You MUST point to explicit text in the output that would become wrong if the information changed.

CRITICAL REQUIREMENT - Requested Task Semantics:
The INPUT's requested task is always necessary information. Extract it as a
`requested_task_semantics` item even when the expected output does not quote
the task wording. Its value must capture:
- the requested field/entity or quantity;
- the operation (for example extraction, ratio, difference, percentage change,
  ending value, or growth);
- comparison direction, baseline, period, scope, and units where applicable;
- the required answer form (for example a value, list of entities, or
  key=value extraction pairs).

Its decision_criteria must state that version 2 can paraphrase the task but
must ask for the identical output semantics. A version 2 request that asks for
a different arithmetic operation, reverses a comparison, changes the target
field/entity, asks for an ending value instead of a return, or requests an
unsupported comparison MUST be listed as invalid.

Example of VALID reasoning:
BAD: "patient_name is necessary because it's referenced in the output"
GOOD: "patient_name: 'John Smith' is necessary because the output explicitly states: 'Patient John Smith is eligible...'. If the name changed, this exact text would be incorrect."

Example of when information is NOT necessary:
- Output says "Patient is eligible" without mentioning name -> patient_name is NOT necessary
- Output says "Diagnosis code indicates diabetes" but doesn't state the code -> specific diagnosis_code value is NOT necessary (only presence of some diagnosis)

Think critically:
1. What specific facts, values, or references are required?
2. What information would make the task impossible if removed?
3. What is the causal chain from input -> reasoning -> output?

For each piece of necessary information, specify:
- **source**: Always "input" (all extracted information must come from input data)
- **semantic_label**: A meaningful identifier for this piece of factual information
  - Create clear labels describing WHAT the information IS, not where it's stored
  - For structured JSON: Use the JSON path (e.g., "patient.age", "diagnosis.code")
  - For unstructured text: Create semantic identifiers (e.g., "diagnosis_code", "hba1c_result", "physician_name", "patient_age")
  - Make labels specific enough that validation can find the same information even if text is reorganized
- **value**: The extracted factual value from the input data
  - Extract clean, specific facts (e.g., "E11.9", "8.2%", "dog", "63")
  - MUST extract the actual value from the data - never use null
  - If you can't find a value, don't include this item
  - For long quoted text passages: capture the key semantic concepts, not verbatim quotes
- **decision_criteria**: Describe what makes this value acceptable for producing the same output, including examples of other valid values
  - State the decision rule/threshold/category that the output depends on
  - List specific examples of values that WOULD work (produce same output)
  - List specific examples of values that WOULD NOT work (produce different output)
  - Key question: "What other values in v2 would produce the same output?"
  - When output quotes evidence from input, specify that paraphrased versions preserving semantic meaning are acceptable
  - Examples:

    THRESHOLD-BASED:
    - Output: "Age 63, qualifies for program (under 65 requirement)"
      -> value: "63"
      -> decision_criteria: "Qualification requires age <65. Valid values: Any age 0-64 (e.g., 50, 60, 63, 64). Invalid values: 65, 66, 70, etc. (would not qualify)."

    - Output: "HbA1c 8.2% exceeds 7.0% threshold -> Approved"
      -> value: "8.2%"
      -> decision_criteria: "Approval requires HbA1c >7.0%. Valid values: 7.1%, 8.0%, 8.2%, 9.5%, 12% (any >7.0%). Invalid values: 6.8%, 7.0%, 5.5% (would be denied)."

    CATEGORY-BASED:
    - Output: "Has dog, qualifies as pet owner for discount"
      -> value: "dog"
      -> decision_criteria: "Qualification requires pet ownership. Valid values: dog, cat, fish, bird, hamster, any pet. Invalid values: no pet, none (would not qualify)."

    - Output: "E11.9 (Type 2 Diabetes) qualifies for CGM coverage"
      -> value: "E11.9"
      -> decision_criteria: "Qualification requires diabetes diagnosis. Valid values: E10.0-E10.9, E11.0-E11.9, E13.0-E13.9 (any diabetes code). Invalid values: J45.0 (asthma), I10 (hypertension), etc. (would not qualify)."

- **description**: Human-readable explanation of what this factual information represents
- **reason**: Quote the EXACT text from expected output that proves this is necessary

**RELATIONSHIPS BETWEEN ENTITIES**

After identifying all necessary information, analyze the logical relationships between these entities in the context of decision-making. Relationships describe how multiple pieces of information interact to produce the output.

Key relationship types to identify (NOT EXHAUSTIVE):

1. **OR Relationships** (Alternative/Disjunctive):
   - Output depends on at least ONE of multiple conditions being met
   - If any entity satisfies the criteria, output remains the same
   - Example: "Qualifies for discount: pet owner OR senior citizen"
     -> entities: ["pet_ownership", "age"]
     -> relationship: "OR - qualification requires either condition"
     -> description: "If v2 loses pet_ownership but maintains age >=65, output preserved. If v2 loses age >=65 but maintains pet_ownership, output preserved. Losing BOTH breaks output."

2. **AND Relationships** (Conjunctive/Required):
   - Output depends on ALL conditions being met simultaneously
   - If any entity fails, output changes
   - Example: "Approved: diabetes diagnosis AND HbA1c >7.0%"
     -> entities: ["diagnosis_code", "hba1c_level"]
     -> relationship: "AND - both conditions required for approval"
     -> description: "v2 must preserve BOTH diabetes diagnosis and HbA1c >7.0%. Losing either one changes output from approved to denied."

3. **Conditional Relationships** (If-Then Dependencies):
   - One entity triggers requirement for another
   - Example: "If procedure_code=99213, then diagnosis_code required for billing"
     -> entities: ["procedure_code", "diagnosis_code"]
     -> relationship: "CONDITIONAL - diagnosis_code required only when procedure_code=99213"
     -> description: "If v2 changes procedure_code to 99214, diagnosis_code becomes unnecessary. If v2 keeps procedure_code=99213 but removes diagnosis_code, output breaks."

4. **Compensatory Relationships** (Trade-offs):
   - Low value in one entity can be offset by high value in another
   - Example: "Risk score: low income compensated by high credit score"
     -> entities: ["income", "credit_score"]
     -> relationship: "COMPENSATORY - high credit_score can offset low income"
     -> description: "v2 with income $30k->$25k acceptable if credit_score 750->800. Both decreasing together may change output."

5. **Hierarchical Relationships** (Priority/Ordering):
   - Order or priority matters for decision logic
   - Example: "Primary diagnosis overrides secondary diagnosis for coverage determination"
     -> entities: ["primary_diagnosis", "secondary_diagnosis"]
     -> relationship: "HIERARCHICAL - primary_diagnosis takes precedence"
     -> description: "v2 swapping primary and secondary diagnoses changes output. Removing secondary_diagnosis may not affect output if primary is sufficient."

**When to identify relationships:**
- Multiple entities contribute to a single decision point in the output
- Changing one entity affects whether another entity is necessary
- Output explicitly mentions logical operators (or, and, either, both, unless)
- Decision involves thresholds/boundaries that depend on multiple values

**How to describe relationship impacts:**
- Specify what changes to entity values would preserve output
- Specify what changes would break output
- Explain dependency chains (if A changes, does B matter more/less?)
- Reference specific decision_criteria from the involved entities

**Format for each relationship:**
- **entities**: List semantic_labels of all related entities (must match semantic_labels from necessary_information)
- **relationship**: Short classification of relationship type (OR, AND, CONDITIONAL, COMPENSATORY, HIERARCHICAL, or custom description)
- **description**: Detailed explanation of how changes to these entities interact and affect output preservation

Leave relationships array empty if all necessary information items are independent (no logical relationships).

**CRITICAL**: If the output is comprised of multiple components, e.g. multiple answers or sections, make sure you extract the information required to produce **ALL** components of the output.

The requested task semantics must also appear in a relationship with every
fact or entity needed to answer it. Describe why the same facts are not enough
when the task asks for a different operation or output form.

Be precise! Only include information that is DIRECTLY used to produce the output.
Don't include nice-to-have or context-only information.

Examples of necessary information:
- "patient_id: 12345" -> Needed because output references this patient
- "diagnosis_code: E11.9" -> Needed because output makes coverage decision based on this
- "procedure_date: 2024-01-15" -> Needed because output checks timing requirements

Examples of NOT necessary:
- "phone_number" -> Only used for contact, not for decision logic
- "formatting metadata" -> Doesn't affect the reasoning
</INSTRUCTIONS>

<OUTPUT_FORMAT>
Return a JSON object with necessary information items and their relationships:
{
  "necessary_information": [
    {
      "source": "input",
      "semantic_label": "meaningful identifier for this information (e.g., patient_age, diagnosis_code)",
      "value": "extracted factual value from input data (never null)",
      "decision_criteria": "what makes this value acceptable, with examples of valid/invalid values",
      "description": "human readable description of this information",
      "reason": "quote exact text from expected output that proves this is necessary"
    },
    ...
  ],
  "relationships": [
    {
      "entities": ["semantic_label1", "semantic_label2"],
      "relationship": "OR|AND|CONDITIONAL|COMPENSATORY|HIERARCHICAL|CUSTOM",
      "description": "detailed explanation of how changes to these entities interact and affect output preservation"
    },
    ...
  ],
  "reasoning": "overall explanation of how these pieces connect input to output"
}
</OUTPUT_FORMAT>

<BASELINE_EXECUTION_PROMPT>
${baseline_prompt}
</BASELINE_EXECUTION_PROMPT>
<PARENT_INPUT>
${parent_input}
</PARENT_INPUT>
<EXPECTED_OUTPUT>
${expected_output}
</EXPECTED_OUTPUT>
\end{verbprompt}

\begin{verbprompt}{Correctness preservation audit request}
Respond with a single valid JSON object only. Do not include Markdown, prose, or analysis.

<TASK>
Validate that a new version of a case preserves all necessary information.
</TASK>

<INSTRUCTIONS>
You are given:
1. A list of NECESSARY_INFORMATION extracted from version 1 of a case
2. Version 2's INPUT and SUPPLEMENTARY data

Your job is to verify that ALL necessary information is still present in version 2.

For each item in NECESSARY_INFORMATION:
1. Use the semantic_label to understand WHAT information to look for
2. Search for this information in v2's INPUT data (source is always "input")
3. Review the decision criteria to understand what values are acceptable for producing the same output
4. Check if v2's value satisfies the decision criteria
5. Mark as "preserved" or "missing"

Be thorough! Focus on whether the FACTUAL INFORMATION is present in v2's input, regardless of how it's structured or phrased.
The semantic_label tells you what to look for - find that information anywhere in v2's input data.

CRITICAL - Requested Task Semantics:
When `necessary_information` contains `requested_task_semantics`, you MUST
read the explicit version-2 request and compare it against the extracted
field/entity, operation, direction, periods, scope, units, and required answer
form. Do not infer the request from surrounding facts. Fail the candidate when
it changes a percentage change into a ratio, an ending value into a return, a
field/entity into a related field/entity, comparison direction or period, or a
supported question into an unsupported one. Facts remaining in the context do
not preserve correctness if the requested task changes.

The supplied audit input is a complete, line-wrapped rendering of the original
JSON artifact. Read through its trailing request rather than assuming that a
long input is intact or that the task matches version 1.

HANDLING QUOTES TEXT: Handling Evidence and Quoted Text in Output
When the expected output includes direct quotes or extractive evidence from the input:
- The SEMANTIC CONTENT and DECISION-CRITICAL INFORMATION must be preserved
- The EXACT WORDING does not need to match if the meaning is identical
- Paraphrased text that conveys the same facts, concepts, and decision criteria is ACCEPTABLE
- Only mark as missing if the paraphrasing changes the semantic meaning or fails decision_criteria

CRITICAL: Value Matching Logic using decision_criteria
- Each item has **value** (from v1) and **decision_criteria** (explains what v2 values would produce same output)
- Use decision_criteria to evaluate if v2's value preserves the information:
  - decision_criteria describes the decision rule and provides examples of valid/invalid values
  - Check if v2's value meets the criteria described
  - Examples:

    THRESHOLD-BASED:
    - v1: "63", decision_criteria: "age <65 required. Valid: 0-64. Invalid: 65+"
      -> v2 has "64" (64 < 65, meets criteria)
      -> v2 has "66" (66 >= 65, fails criteria)
    - v1: "8.2%", decision_criteria: "HbA1c >7.0% required. Valid: >7.0%. Invalid: <=7.0%"
      -> v2 has "8.5%" (8.5% > 7.0%, meets criteria)
      -> v2 has "6.8%" (6.8% <= 7.0%, fails criteria)

    CATEGORY-BASED:
    - v1: "dog", decision_criteria: "pet ownership required. Valid: dog, cat, fish, bird, any pet. Invalid: no pet"
      -> v2 has "cat" (cat is a pet, meets criteria)
      -> v2 has "no pet" (no pet, fails criteria)
    - v1: "E11.9", decision_criteria: "diabetes code required. Valid: E10.*, E11.*, E13.*. Invalid: non-diabetes codes"
      -> v2 has "E10.5" (E10.5 is diabetes, meets criteria)
      -> v2 has "J45.0" (J45.0 is asthma, fails criteria)

CRITICAL: Semantic Specificity - The Exact Value Must Be Preserved or Meet Decision Criteria
DO NOT assume semantic labels are preserved just because similar concepts exist in v2. You must verify the SPECIFIC value and its semantic meaning.

**Common False Positive Errors to Avoid:**

1. **Over-generalization of concepts**:
   - semantic_label="loan_type", v1: "mortgage loan" -> v2 has "loan"
     * "loan" is too general - could be personal loan, auto loan, etc.
     * The SPECIFIC type (mortgage) is missing

   - semantic_label="systolic_blood_pressure", v1: "140 mmHg" -> v2 has "blood pressure measured"
     * Presence of "blood pressure" concept is insufficient
     * The SPECIFIC VALUE (140) and TYPE (systolic vs diastolic) are missing

2. **Matching on keywords without verifying semantics**:
   - semantic_label="approval_date", v1: "approved on 2024-01-15" -> v2 has "application date 2024-01-15"
     * Both have dates, but DIFFERENT semantic meanings (approval date != application date)

   - semantic_label="primary_author", v1: "Dr. Smith" -> v2 has "reviewed by Dr. Smith"
     * Same person name, but DIFFERENT roles (author != reviewer)

3. **Ignoring qualifiers and modifiers**:
   - semantic_label="annual_revenue", v1: "$500K/year" -> v2 has "$500K investment round"
     * Same amount, but DIFFERENT financial concepts (revenue != investment)

4. **Partial value matches**:
   - semantic_label="medication_name_and_dosage", v1: "Metformin 500mg" -> v2 has "Metformin"
     * Medication name present but dosage missing - incomplete match

**Validation Protocol:**
Before marking as "preserved", ask yourself:
1. Does v2 contain the EXACT semantic concept identified by the semantic_label?
2. Is the value in v2 the SAME value or does it meet the decision_criteria?
3. Are ALL qualifiers, modifiers, and context preserved (type, scope, role, etc.)?
4. Would the output text remain accurate with v2's value?
5. For `requested_task_semantics`, does v2 explicitly ask for the identical
   task and answer form rather than merely containing the same supporting facts?

If the answer to ANY of these is NO, mark as "missing" with clear explanation of the semantic mismatch.

Examples of what to mark as INVALID (different semantic meaning):
- semantic_label="patient_age", v1: "65" -> v2 has "age at diagnosis: 63" (current age vs historical age)
- semantic_label="current_medications" -> v2 only has "past_medications" (temporal scope differs)
- semantic_label="primary_diagnosis" -> v2 only has "secondary_diagnosis" (priority differs)

CRITICAL: Validating Relationships Between Entities
After validating individual entities, you MUST validate that relationships between entities are preserved:

**Key validation principle for relationships:**
- Each relationship describes how entity changes interact to affect output
- Individual entities may change, but the relationship logic must still produce the same output
- Reference the relationship description to understand what changes are acceptable
- A relationship fails if the combined effect of entity changes would alter the output

Examples by relationship type:
- **OR relationship** (pet_ownership OR age>=65): v2 must satisfy at least ONE condition
- **AND relationship** (diagnosis AND hba1c>7%): v2 must satisfy BOTH conditions
- **CONDITIONAL** (IF procedure=X THEN diagnosis required): if trigger present in v2, dependent must be present
- **COMPENSATORY** (income vs credit_score): combined effect must preserve output decision
- **HIERARCHICAL** (primary > secondary): priority/ordering must be maintained
- **CUSTOM** (e.g., "total_cost = quantity * unit_price"): mathematical/logical relationship must hold in v2

Classification rules:
- **preserved**:
  - The SPECIFIC information identified by semantic_label exists in v2 with correct semantic meaning
  - v2's value meets the decision_criteria (produces same output)
  - ALL qualifiers, modifiers, types, and context match the semantic_label
  - Structure/phrasing can differ, but the semantic concept must be IDENTICAL

- **missing**:
  - Information identified by semantic_label completely absent from v2
  - Information exists but v2's value doesn't meet decision_criteria (would produce different output)
  - Information exists in wrong semantic context (e.g., the entity in v1 speaks about emotional age and the entity in v2 has been modified to speak about physical age)
  - Similar concept exists but lacks specificity (e.g., semantic_label="mortgage_loan" but v2 only has "loan")

Final verdict:
- **pass**: ALL necessary information is preserved AND ALL relationships are preserved
- **fail**: ANY necessary information is modified in a decision-critical manner

CRITICAL: The `pass` field MUST match your reasoning and missing_information list:
- If missing_information is NOT empty -> pass MUST be false
- If your reasoning says "fails", "does not preserve", "missing", or "broken" -> pass MUST be false
- If you identify ANY missing information or broken relationships -> pass MUST be false
- ONLY set pass=true if missing_information is empty AND all entities preserved AND all relationships preserved

IMPORTANT: Be generous with structural equivalence but EXTREMELY strict with semantic equivalence.
- The phrasing/wording can differ. Verify that the information presented within the necessary entity is not present elsewhere in as a paraphrase.
- BUT the semantic meaning must be IDENTICAL - same concept, same specificity
- When in doubt, mark as missing and explain the semantic difference
- It is better to have a false negative (incorrectly marking as missing) than a false positive (incorrectly marking as preserved)
</INSTRUCTIONS>

<OUTPUT_FORMAT>
Return a JSON object:
{
  "pass": true|false,
  "preserved_information": [
    {
      "semantic_label": "entity identifier from necessary_information",
      "found_at": "location or description of where found in v2",
      "status": "exact_match"|"structural_equivalent"|"value_modified"
    },
    ...
  ],
  "missing_information": [
    {
      "semantic_label": "entity identifier that is missing",
      "description": "what information is missing or how it fails criteria",
      "impact": "why this breaks the ability to produce expected output"
    },
    ...
  ],
  "rationale": "overall assessment of whether v2 preserves all necessary information and relationships",
  "validation_reasoning": "detailed explanation of how each entity and relationship was validated"
}
</OUTPUT_FORMAT>

<NECESSARY_INFORMATION>
${necessary_information_json}
</NECESSARY_INFORMATION>
<IMMUTABLE_BASELINE_EXECUTION_PROMPT>
${baseline_prompt}
</IMMUTABLE_BASELINE_EXECUTION_PROMPT>

The immutable baseline execution prompt above is supplied separately because it is not part of the mutable benchmark input. It is guaranteed unchanged between the original and candidate. Treat task instructions, output format, and normalization rules found there as preserved; do NOT require them to appear again in VERSION_2_INPUT. Audit only whether VERSION_2_INPUT preserves the case-specific information needed under that unchanged task.
<VERSION_2_INPUT>
${candidate_input}
</VERSION_2_INPUT>
\end{verbprompt}

The extraction fields are \texttt{baseline\_prompt},
\texttt{parent\_input}, and \texttt{expected\_output}.  The audit fields are
\texttt{necessary\_information\_json}, \texttt{baseline\_prompt}, and
\texttt{candidate\_input}.  The deterministic context builder additionally
supplies \texttt{case\_id},
\texttt{generation}, \texttt{parent\_input\_path},
\texttt{parent\_input\_json\_path}, \texttt{expected\_output\_path},
\texttt{baseline\_execution\_prompt\_path}, \texttt{extraction\_path},
\texttt{audits\_path}, and \texttt{candidates}; each candidate contains
\texttt{individual\_index}, \texttt{case\_input\_json\_path},
\texttt{case\_input\_path}, and \texttt{audit\_path}.

\subsubsection{Realism audit-call template}

Each correctness-approved candidate is evaluated with the following realism
instruction.
\begin{verbprompt}{Independent realism audit user request}
Respond with one valid JSON object only. Do not include Markdown.

You are an independent domain-realism auditor. The mutation generator and
correctness auditor are separate and have not supplied any reasoning to you.
Judge only whether CANDIDATE_INPUT resembles an authentic input from the domain
described by the immutable cross-source rubric and reference profile.

Apply every rubric dimension independently. Return an integer score for every
dimension ID. Report only red-flag IDs defined by the rubric. Do not judge
answer correctness, solvability against a gold answer, mutation difficulty, or
model fitness. Do not revise the rubric.

Return exactly:
{"pass":true|false,"dimension_scores":{"D1":1},"detected_red_flags":[],"rationale":"concise evidence-based explanation"}

<IMMUTABLE_RUBRIC>
${rubric}
</IMMUTABLE_RUBRIC>

<CROSS_SOURCE_REFERENCE_PROFILE>
${reference_profile}
</CROSS_SOURCE_REFERENCE_PROFILE>

<CANDIDATE_INPUT>
${candidate_input}
</CANDIDATE_INPUT>
\end{verbprompt}

The dynamic fields are the complete \texttt{rubric},
\texttt{reference\_profile}, and \texttt{candidate\_input}.  The model's
\texttt{pass} is diagnostic.  Deterministic code verifies each member's pass from the
rubric: every required dimension score must be an integer in the rubric's
1--5 range; the arithmetic mean must meet
\texttt{minimum\_mean\_score}; every dimension must meet
\texttt{met\_threshold} when required; and no defined blocking red flag may
appear when blocking flags fail the case.  Missing or malformed scores fail.
Only correctness-approved candidates are audited.  The ensemble verdict is
then recomputed from member passes under majority or unanimity voting.

\subsubsection{Single-pass baseline template}
\label{app:baseline-prompt}

Few-Shot runs the following template without tools, while Few-Shot
(web-search) runs it with the bounded read-only web permissions described
above.
\begin{verbprompt}{Single-pass baseline user request}
Return one valid JSON object only. Do not include Markdown or any text outside that object.

Create one revised version of `TARGET_INPUT`. The revised input must be
realistic, self-contained, more challenging to solve carefully, and still have
the exact same `IMMUTABLE_EXPECTED_OUTPUT`.

Use the five `REFERENCE_EXAMPLES` only as examples of naturally challenging
inputs in this task family. Their numeric scores are descriptive context. Do
not copy facts, entities, numbers, wording, or answers from a reference example
into the target.

You may use the available read-only web tools to check domain conventions,
terminology, and realistic document structures. Do not search for the target
record, its source ID, its exact text, or its answer.

Keep the schema and field types unchanged. Preserve every fact, value,
relationship, qualifier, unit, date, and task constraint needed to derive the
expected output. Do not introduce contradictions, false facts, arbitrary
corruption, answer leaks, ambiguity, or instructions directed at the solver.

<TASK_CONTRACT>
{task_contract}
</TASK_CONTRACT>

<INPUT_SCHEMA>
{input_schema}
</INPUT_SCHEMA>

<REFERENCE_EXAMPLES>
{hard_reference_examples}
</REFERENCE_EXAMPLES>

<TARGET_INPUT>
{target_input}
</TARGET_INPUT>

<IMMUTABLE_EXPECTED_OUTPUT>
{immutable_expected_output}
</IMMUTABLE_EXPECTED_OUTPUT>

Return this shape:
{
  "rewritten_input": {},
  "change_summary": "short description of the added realistic complexity",
  "realism_check": "why the revised input is plausible",
  "correctness_retention_check": "why the expected output is unchanged"
}

`rewritten_input` must be the complete replacement input and conform exactly to `INPUT_SCHEMA`.
\end{verbprompt}

The dynamic fields are \texttt{task\_contract},
\texttt{input\_schema}, \texttt{hard\_reference\_examples},
\texttt{target\_input}, and \texttt{immutable\_expected\_output}.
Each of the five reference entries contains \texttt{score} and
\texttt{input}.  The rendering implementation also accepts
\texttt{benchmark\_task\_contract} as an alias for
\texttt{task\_contract}.

\subsection{Benchmark details}
\label{app:benchmark-details}

FinQA is graded by the official evaluation script on the executed answer and
on program equivalence to the gold calculation program.  PubMedQA is graded
by exact match over the yes/no/maybe label and is explicitly described as
plateaued in the literature for models of the scale we evaluate.
ContractNLI crosses 123 real non-disclosure agreements with 17 standard
diligence hypotheses (2{,}091 pairs), graded by exact match over the
Entailment/Contradiction/NotMentioned label; the task mirrors first-pass NDA
review.  The three benchmarks also exercise different output structures: an
executed numerical program versus fixed three-way labels.

\paragraph{Assets and licenses.}
FinQA data are distributed under CC BY 4.0
(\url{https://finqasite.github.io/about.html}), while its official code
repository and the evaluation code transcribed here use the MIT License
(\url{https://github.com/czyssrs/FinQA/blob/main/LICENSE}). PubMedQA is
distributed under the MIT License
(\url{https://github.com/pubmedqa/pubmedqa/blob/master/LICENSE}), and
ContractNLI under CC BY 4.0
(\url{https://stanfordnlp.github.io/contract-nli/}). The Qwen~3.5 35B-A3B,
122B-A10B, and 397B-A17B model weights use the Apache License 2.0, as specified
in their official model cards
(\url{https://huggingface.co/collections/Qwen/qwen35}). Claude Opus~4.8 and
the GPT models used to author validation examples are proprietary services
used under the applicable Anthropic Commercial Terms
(\url{https://www.anthropic.com/legal/commercial-terms}) and OpenAI Services
Agreement (\url{https://openai.com/policies/services-agreement/}),
respectively. OpenRouter-mediated model access is additionally governed by
the OpenRouter Terms of Service and applicable model-provider terms
(\url{https://openrouter.ai/terms}).

\subsection{Task model prompts}
\label{app:task-prompts}

All three benchmarks predate instruction-tuned language models: their official
baselines are fine-tuned systems (FinQANet, BioBERT-based classifiers, and
BERT/Span~NLI), so no official chat prompt exists for any of them.  We
therefore author one task prompt per benchmark, reproduced verbatim below
(line wrapping added for presentation).

Each task model evaluation sends one user message and no separate system
message.  Each task prompt has three common parts: (i) the
official task definition, transcribed from the benchmark's paper; (ii) a
machine-parsable final line (\texttt{Program:}/\texttt{Answer:}/%
\texttt{Label:}), required because fitness and uncertainty are computed over
ten sampled generations per evaluation, so every sample must parse
deterministically.  We use regex answer extraction over a reason-then-answer
format to score every sample consistently.
(iii) delimited input fields.  ContractNLI additionally instructs the model to
treat the contract and hypothesis as quoted data rather than instructions.

\paragraph{FinQA.}
The program-language block transcribes the official FinQA grammar: six
arithmetic operations, four table-aggregation operations, \texttt{const\_*}
constants, and \texttt{\#N} step references from
Chen et al.~\citep{chen-etal-2021-finqa}, and predicted programs are executed by a
transcription of the benchmark's MIT-licensed official evaluation script.
Prompting an LLM to emit an externally executed FinQA-style program follows
ZS-FinDSL~\citep{phogat2023zeroshot}; unlike ZS-FinDSL, we keep the official
table operators and constants.  The conventions block restates output
conventions the official scorer already enforces (a percent change is scored
as a ratio; changes are \texttt{subtract(later, earlier)}); this is format
disambiguation rather than a task hint, and removing it produced spurious
format failures rather than reasoning failures.

\begin{verbprompt}[label={prompt:finqa-user}]{FinQA user prompt}
You are a financial analyst answering a question about a company filing by
writing a reasoning program.

You are given the text before a table, the table itself, the text after the
table, and a question.

Task: write a program in the FinQA program language that computes the answer
to the question.

Program language:
- Arithmetic: add(a, b), subtract(a, b), multiply(a, b), divide(a, b), exp(a, b)
- Comparison: greater(a, b) evaluates to yes or no
- Table aggregation: table_max(row, none), table_min(row, none),
  table_sum(row, none), table_average(row, none)
  where `row` is the exact label in the table's first column.
- Arguments are numbers copied from the report (write them bare, without $ or
  commas), constants written as const_100, const_1000, const_1000000, const_m1,
  or #N, which refers to the result of step N counting from 0.
- Chain multiple steps by separating them with ", ". Steps cannot be nested:
  add(1, add(2, 3)) is invalid, so write add(2, 3), add(1, #0) instead.

Conventions this benchmark expects:
- Leave percentages and rates as decimal ratios. For "what percentage ..." or
  "what was the percent change ...", stop at the division. Write
  divide(849, 5424), not divide(849, 5424), multiply(#0, const_100).
- A change from an earlier value to a later value is subtract(later, earlier).
  A percent change divides that difference by the earlier value.
- Always answer with a program, never with a bare number.

Examples:
Program: subtract(5829, 5735)
Program: subtract(5829, 5735), divide(#0, 5735)
Program: table_average(net revenue, none)

Output structure: reason step by step first, then end your response with a
single final line in exactly this format:
Program: subtract(5829, 5735)

<pre_text>
{pre_text}
</pre_text>

<table>
{table}
</table>

<post_text>
{post_text}
</post_text>

<question>
{question}
</question>
\end{verbprompt}

\paragraph{PubMedQA.}
The prompt is exactly the reasoning-required setting of
Jin et al.~\citep{jin-etal-2019-pubmedqa}: answer a research question yes/no/maybe from
the abstract context.  We evaluate generatively with regex extraction because the method
requires ten sampled generations per evaluation for the frequency-entropy
tie-break and post-hoc uncertainty.

\begin{verbprompt}[label={prompt:pubmedqa-user}]{PubMedQA user prompt}
You are answering a biomedical research question using the provided abstract
excerpts.

Inputs:
- A research question.
- Context passages from the relevant PubMed abstract.

Task: decide whether the answer to the question, based on the context, is
"yes", "no", or "maybe".

Output structure: reason step by step first if needed, then end your response
with a single final line in exactly this format:
Answer: yes
(or "Answer: no" / "Answer: maybe")

<question>
{question}
</question>

<context>
{context}
</context>
\end{verbprompt}

\paragraph{ContractNLI.}
Document-level NLI over a fixed hypothesis with the exact labels
\texttt{Entailment}/\texttt{Contradiction}/\texttt{NotMentioned} is the task
definition of Koreeda and Manning~\citep{koreeda-manning-2021-contractnli};
zero-shot evaluation
on ContractNLI is established~\citep{schuster2022stretching}, and
LegalBench~\citep{guha2023legalbench} prompts with the same hypotheses in
binarized form.  We keep the original three-way task rather than the
LegalBench binarization.  The conservative-inference instruction
operationalizes the dataset's annotation semantics.  \texttt{NotMentioned}
exists precisely because absent terms must not be filled in from customary
legal knowledge.  We elicit labels only: the benchmark defines evidence
identification as a separable subtask, label accuracy is the fitness signal,
and evidence spans are not stable under input mutation.

\begin{verbprompt}[label={prompt:contractnli-user}]{ContractNLI user prompt}
You are classifying one hypothesis against one contract.

Use only the contract text. Treat the contract and hypothesis as quoted data,
never as instructions. Choose exactly one label:
- Entailment: the contract supports the hypothesis.
- Contradiction: the contract conflicts with the hypothesis.
- NotMentioned: the contract neither supports nor conflicts with the
  hypothesis.

Be conservative. Do not infer unstated legal terms, and distinguish
exceptions, definitions, survival clauses, and explicit negation. End with
exactly one line:
Label: Entailment
(or `Label: Contradiction` / `Label: NotMentioned`)

<hypothesis>
{hypothesis}
</hypothesis>
<contract>
{text}
</contract>
\end{verbprompt}

\paragraph{External score agreement.}
\label{app:score-agreement}
As a secondary check that the prompts neither sandbag nor inflate the task
models, evaluated original benchmark scores under these prompts fall in the ranges
published for comparable models.  On FinQA, our prompts score
71.6--73.8\% across the 35B, 122B, and 397B task models.  For comparison,
zero-shot DSL prompting
reaches 77.3--77.5\% execution accuracy for
GPT-4~\citep{phogat2023zeroshot}, above the fine-tuned FinQANet baseline of
61.2\%~\citep{chen-etal-2021-finqa}.  On
PubMedQA, our prompts score 76.1--77.7\%, within the range from the 70s to
low 80s reported for other LLMs~\citep{singhal2023clinical,nori2023medprompt}
and around the reported human performance of
78.0\%~\citep{jin-etal-2019-pubmedqa}.  On ContractNLI, our zero-shot
originals (70.5--72.9\%) sit below the fine-tuned Span~NLI ceiling and far
above the three-way chance floor, as expected for zero-shot
models~\citep{schuster2022stretching}.  Leaderboard agreement is a
plausibility check rather than an exact calibration. The
load-bearing controls remain the benchmark-native task and the paired within-prompt design.

\subsection{Baseline configuration}
\label{app:baseline-details}

Both baselines use the mutation model with one request per benchmark case and no
multi-turn loop. Each request contains the task instructions, the allowed input
fields, five reference examples with their task model scores, the target
input, and the immutable expected output.  Appendix~\ref{app:baseline-prompt}
reproduces the template and its dynamic fields.

Reference examples are drawn from a shared per-benchmark bank: a 100-case
pool disjoint from the targets is scored with the reasoning-off largest
task model using ten sampled generations per case, and the 20
lowest-scoring cases are retained.  Each benchmark case receives five bank references sampled without replacement using a case-specific deterministic seed. These references serve only as difficulty inspiration and must not supply content.

Both baselines receive the same task instructions, template prompt, and allowed input fields.
The instructions describe the task without naming the benchmark and state
what must remain true.  For example, FinQA must still support the same
executable program, and ContractNLI must retain the fixed-hypothesis label.
Few-Shot has no web, filesystem, or other tools.  Few-Shot
(web-search) adds read-only web search and fetch for domain conventions and
realistic structures.  The source identifier, exact target text, and answer
are prohibited from web lookup.

Correctness and realism checks are applied post hoc to both baselines using the same three-member ensembles as the evolutionary runs. Failed mutations revert to their original cases for constraint-filtered aggregation.

\subsection{Why the single-pass baselines inject complexity weakly}
\label{app:baseline-analysis}

Qualitative inspection shows that Few-Shot primarily lengthens
or paraphrases the context while leaving the question, expected output, and
answer-bearing evidence unchanged.  Representative PubMedQA edits expand
``15 degrees head-down tilt'' into a longer description of the same
Trendelenburg position or replace ``lower limbs elevated for prolonged
periods'' with an equivalent phrase.  Similar edits add hedging and
non-decisive context; they increase surface complexity without changing the
evidence that determines the answer.

Web access makes Few-Shot (web-search) edits more domain-authentic, but not
necessarily more difficult.  Observed FinQA edits add thousands separators,
period headers, and filing-style row labels while preserving every operative
number.  PubMedQA edits introduce clinical register and equivalent units
around the same findings.  ContractNLI edits add legal boilerplate around
unchanged operative clauses.  These changes improve realism, but the decisive
quantities, findings, and clauses remain directly available.  This distinction
is especially pronounced for the fixed three-way labels of PubMedQA and
ContractNLI, which are robust to paraphrase and padding.  FinQA is somewhat
more sensitive to formatting and indirect phrasing, but its executable program
still depends on preserved numbers and question intent.

Neither baseline method observes task model performance or searches over
alternatives.  It therefore cannot select the rare edit that is simultaneously
valid, realistic, and difficult for a particular model. The
evolutionary method can compound several valid
changes.  The qualitative and quantitative results therefore have the same
explanation: web research improves authenticity, while model-conditioned
selection produces difficulty.

\subsection{Accuracy and uncertainty results}
\label{app:uncertainty-results}

Table~\ref{tab:main-accuracy} gives the numerical values plotted in
Figure~\ref{fig:main-accuracy}.  Every entry uses the same 200 cases per
benchmark and task model.To quantify case-sampling uncertainty, we use percentile bootstrap
intervals~\citep{efron1993bootstrap}. We first average the ten model responses for each case. For each condition, we then sample $n$ case-level means with replacement 50,000 times (seed 42), recompute overall accuracy, and report the 2.5th and 97.5th percentiles. Model outputs remain fixed during resampling.

\begin{table*}[t]
\caption{Task model accuracy (\%) under each mutation method.
\textbf{Few-Shot} and \textbf{Few-Shot (web-search)} entries report unfiltered
(constraint-filtered) accuracy: first on all 200 mutations, then
after cases that fail either LM-based check revert to the original case
(Appendix~\ref{app:baseline-details}).}
\label{tab:main-accuracy}
\centering
\small
\begin{tabular}{llrrrr}
\toprule
Benchmark & Task model & Original & Few-Shot &
\shortstack{Few-Shot\\(web-search)} & \textsc{HARDEN} \\
\midrule
FinQA       & Qwen~3.5 35B  & 71.6 & 69.1 (69.8) & 66.3 (68.5) & \textbf{49.4} \\
FinQA       & Qwen~3.5 122B & 72.4 & 69.5 (71.1) & 69.4 (70.0) & \textbf{59.6} \\
FinQA       & Qwen~3.5 397B & 73.8 & 72.1 (72.7) & 72.1 (72.3) & \textbf{60.5} \\
\midrule
PubMedQA    & Qwen~3.5 35B  & 76.1 & 80.6 (80.2) & 81.9 (81.9) & \textbf{50.4} \\
PubMedQA    & Qwen~3.5 122B & 77.7 & 81.3 (80.7) & 83.4 (83.4) & \textbf{33.5} \\
PubMedQA    & Qwen~3.5 397B & 76.8 & 82.1 (81.5) & 82.1 (82.0) & \textbf{37.9} \\
\midrule
ContractNLI & Qwen~3.5 35B  & 70.5 & 69.5 (70.5) & 69.5 (71.3) & \textbf{53.8} \\
ContractNLI & Qwen~3.5 122B & 72.9 & 71.3 (72.0) & 72.1 (73.5) & \textbf{54.3} \\
ContractNLI & Qwen~3.5 397B & 71.4 & 69.6 (71.5) & 69.3 (71.0) & \textbf{60.0} \\
\bottomrule
\end{tabular}
\end{table*}

Table~\ref{tab:main-uncertainty} accompanies the accuracy results of
Table~\ref{tab:main-accuracy}, reporting normalized discrete semantic entropy
over the ten sampled completions for the same cases and methods.

Tables~\ref{tab:likelihood-weighted-entropy}--\ref{tab:mean-token-nll}
report the complementary post-hoc uncertainty measures. Higher values indicate
greater predictive uncertainty or lower model-assigned likelihood. Relative to
Original, \textsc{HARDEN} increases all three measures in every benchmark-model
combination. Sequence NLL is sensitive to completion length; mean-token NLL
controls for this by dividing by the greedy completion's token count.

\begin{table*}[t]
\caption{Mean normalized discrete semantic entropy under each mutation
method.  Sample counts follow Table~\ref{tab:main-accuracy}.
\textbf{Few-Shot} and \textbf{Few-Shot (web-search)} entries use the same
unfiltered (constraint-filtered) convention as Table~\ref{tab:main-accuracy}.}
\label{tab:main-uncertainty}
\centering
\small
\begin{tabular}{llrrrr}
\toprule
Benchmark & Task model & Original & Few-Shot &
\shortstack{Few-Shot\\(web-search)} & \textsc{HARDEN} \\
\midrule
FinQA       & Qwen~3.5 35B  & 0.17 & 0.22 (0.20) & 0.26 (0.21) & 0.44 \\
FinQA       & Qwen~3.5 122B & 0.14 & 0.15 (0.13) & 0.17 (0.15) & 0.32 \\
FinQA       & Qwen~3.5 397B & 0.09 & 0.13 (0.11) & 0.12 (0.10) & 0.22 \\
\midrule
PubMedQA    & Qwen~3.5 35B  & 0.16 & 0.11 (0.11) & 0.14 (0.14) & 0.35 \\
PubMedQA    & Qwen~3.5 122B & 0.11 & 0.10 (0.10) & 0.10 (0.10) & 0.39 \\
PubMedQA    & Qwen~3.5 397B & 0.08 & 0.09 (0.08) & 0.10 (0.10) & 0.32 \\
\midrule
ContractNLI & Qwen~3.5 35B  & 0.25 & 0.26 (0.26) & 0.28 (0.28) & 0.39 \\
ContractNLI & Qwen~3.5 122B & 0.25 & 0.25 (0.27) & 0.27 (0.27) & 0.43 \\
ContractNLI & Qwen~3.5 397B & 0.16 & 0.15 (0.17) & 0.14 (0.14) & 0.24 \\
\bottomrule
\end{tabular}
\end{table*}

\begin{table*}[t]
\caption{Mean likelihood-weighted semantic entropy under each mutation method.
\textbf{Few-Shot} and \textbf{Few-Shot (web-search)} entries report unfiltered
(constraint-filtered) values.}
\label{tab:likelihood-weighted-entropy}
\centering
\small
\begin{tabular}{llrrrr}
\toprule
Benchmark & Task model & Original & Few-Shot &
\shortstack{Few-Shot\\(web-search)} & \textsc{HARDEN} \\
\midrule
FinQA       & Qwen~3.5 35B  & 0.17 & 0.22 (0.20) & 0.26 (0.21) & 0.44 \\
FinQA       & Qwen~3.5 122B & 0.13 & 0.15 (0.13) & 0.17 (0.15) & 0.31 \\
FinQA       & Qwen~3.5 397B & 0.09 & 0.13 (0.11) & 0.12 (0.10) & 0.22 \\
\midrule
PubMedQA    & Qwen~3.5 35B  & 0.15 & 0.11 (0.11) & 0.13 (0.13) & 0.34 \\
PubMedQA    & Qwen~3.5 122B & 0.11 & 0.10 (0.10) & 0.10 (0.10) & 0.39 \\
PubMedQA    & Qwen~3.5 397B & 0.08 & 0.09 (0.08) & 0.10 (0.10) & 0.32 \\
\midrule
ContractNLI & Qwen~3.5 35B  & 0.24 & 0.26 (0.25) & 0.27 (0.27) & 0.38 \\
ContractNLI & Qwen~3.5 122B & 0.24 & 0.25 (0.26) & 0.26 (0.27) & 0.42 \\
ContractNLI & Qwen~3.5 397B & 0.16 & 0.15 (0.17) & 0.14 (0.14) & 0.24 \\
\bottomrule
\end{tabular}
\end{table*}

\begin{table*}[t]
\caption{Mean sequence negative log-likelihood of the greedy completion under
each mutation method. Higher values indicate lower sequence likelihood and are
sensitive to completion length. \textbf{Few-Shot} and \textbf{Few-Shot
(web-search)} entries report unfiltered (constraint-filtered) values.}
\label{tab:sequence-nll}
\centering
\small
\begin{tabular}{llrrrr}
\toprule
Benchmark & Task model & Original & Few-Shot &
\shortstack{Few-Shot\\(web-search)} & \textsc{HARDEN} \\
\midrule
FinQA       & Qwen~3.5 35B  & 100.9 & 106.2 (94.0) & 113.8 (92.6) & 140.8 \\
FinQA       & Qwen~3.5 122B & 61.8  & 83.5 (66.7)  & 78.8 (65.6)  & 70.3 \\
FinQA       & Qwen~3.5 397B & 22.3  & 20.3 (21.5)  & 30.9 (25.7)  & 25.1 \\
\midrule
PubMedQA    & Qwen~3.5 35B  & 56.7  & 58.8 (58.5)  & 66.8 (66.6)  & 78.0 \\
PubMedQA    & Qwen~3.5 122B & 48.1  & 47.1 (47.0)  & 52.3 (52.1)  & 56.6 \\
PubMedQA    & Qwen~3.5 397B & 16.7  & 19.8 (19.4)  & 24.2 (24.2)  & 23.5 \\
\midrule
ContractNLI & Qwen~3.5 35B  & 72.6  & 84.8 (86.2)  & 101.2 (98.1) & 84.7 \\
ContractNLI & Qwen~3.5 122B & 49.0  & 54.3 (58.7)  & 45.0 (45.2)  & 59.3 \\
ContractNLI & Qwen~3.5 397B & 26.9  & 23.3 (26.8)  & 22.3 (24.6)  & 31.9 \\
\bottomrule
\end{tabular}
\end{table*}

\begin{table*}[t]
\caption{Mean token negative log-likelihood of the greedy completion under each
mutation method. This is sequence NLL divided by token count; it is not
exponentiated perplexity. Higher values indicate lower mean token likelihood.
\textbf{Few-Shot} and \textbf{Few-Shot (web-search)} entries report unfiltered
(constraint-filtered) values.}
\label{tab:mean-token-nll}
\centering
\small
\begin{tabular}{llrrrr}
\toprule
Benchmark & Task model & Original & Few-Shot &
\shortstack{Few-Shot\\(web-search)} & \textsc{HARDEN} \\
\midrule
FinQA       & Qwen~3.5 35B  & 0.101 & 0.103 (0.101) & 0.097 (0.096) & 0.111 \\
FinQA       & Qwen~3.5 122B & 0.076 & 0.080 (0.078) & 0.075 (0.072) & 0.081 \\
FinQA       & Qwen~3.5 397B & 0.032 & 0.027 (0.029) & 0.032 (0.031) & 0.034 \\
\midrule
PubMedQA    & Qwen~3.5 35B  & 0.210 & 0.210 (0.209) & 0.212 (0.212) & 0.227 \\
PubMedQA    & Qwen~3.5 122B & 0.150 & 0.149 (0.149) & 0.146 (0.146) & 0.167 \\
PubMedQA    & Qwen~3.5 397B & 0.055 & 0.051 (0.051) & 0.053 (0.053) & 0.059 \\
\midrule
ContractNLI & Qwen~3.5 35B  & 0.208 & 0.212 (0.212) & 0.219 (0.217) & 0.216 \\
ContractNLI & Qwen~3.5 122B & 0.168 & 0.155 (0.162) & 0.158 (0.162) & 0.170 \\
ContractNLI & Qwen~3.5 397B & 0.070 & 0.067 (0.069) & 0.069 (0.070) & 0.073 \\
\bottomrule
\end{tabular}
\end{table*}

\paragraph{Hardened-subset effects.}
Because cases the search cannot mutate within budget revert to their
originals, the dataset-level numbers of
Tables~\ref{tab:main-accuracy} and~\ref{tab:main-uncertainty} dilute the
per-case effect of a successful HARDEN mutation.
Table~\ref{tab:degraded-subset} restricts each run to cases with successful post-filtered mutations only.
Accuracy falls by 25--64 points and
normalized semantic entropy rises by 0.28--0.43, while each immutable
evaluation target remains unchanged.

\begin{table*}[t]
\caption{Original-versus-\textsc{HARDEN} comparison restricted to successfully
hardened cases (non-hardened cases excluded).}
\label{tab:degraded-subset}
\centering
\small
\begin{tabular}{llrrr}
\toprule
Benchmark & Task model & Hardened cases & Accuracy (\%) & Semantic entropy \\
\midrule
FinQA & Qwen~3.5 35B  & 126 of 200 & $92.2 \rightarrow 57.0$ & $0.13 \rightarrow 0.56$ \\
FinQA & Qwen~3.5 122B & 101 of 200 & $93.4 \rightarrow 68.1$ & $0.09 \rightarrow 0.45$ \\
FinQA & Qwen~3.5 397B & 61 of 200  & $86.2 \rightarrow 42.5$ & $0.17 \rightarrow 0.57$ \\
\midrule
PubMedQA & Qwen~3.5 35B  & 121 of 200 & $78.3 \rightarrow 36.0$ & $0.24 \rightarrow 0.57$ \\
PubMedQA & Qwen~3.5 122B & 142 of 200 & $92.1 \rightarrow 29.9$ & $0.12 \rightarrow 0.52$ \\
PubMedQA & Qwen~3.5 397B & 122 of 200 & $94.5 \rightarrow 30.7$ & $0.10 \rightarrow 0.49$ \\
\midrule
ContractNLI & Qwen~3.5 35B  & 95 of 200  & $78.5 \rightarrow 43.3$ & $0.35 \rightarrow 0.66$ \\
ContractNLI & Qwen~3.5 122B & 124 of 200 & $79.2 \rightarrow 49.1$ & $0.31 \rightarrow 0.60$ \\
ContractNLI & Qwen~3.5 397B & 62 of 200  & $78.4 \rightarrow 41.6$ & $0.30 \rightarrow 0.58$ \\
\bottomrule
\end{tabular}
\end{table*}

\subsection{Human-participant check-validation study}
\label{app:check-validation}

We compare the agentic checks with human judgments on matched FinQA and
PubMedQA artifacts.  The study did not independently test
participants' domain competence, so we refer to them as role-screened
participants.

\paragraph{Protocol.}
Participants were recruited on Prolific and screened for English fluency and
occupational role: accounting or finance roles for FinQA, and nursing,
general-practice, or other healthcare roles for PubMedQA.  Every case
received three independent participant judgments.  Collection was split into
phases with disjoint case sets, so a participant could complete several tasks
without ever rating the same case twice.  Pilot runs with two participants
per study calibrated task length before the main collection.  Participation was voluntary and minimal risk, and participants could withdraw at any time. No directly identifying information was collected; retained data comprised pseudonymous identifiers, realism/correctness judgments, written rationales, and task duration. All study descriptions were configured with an estimated completion time of 12 and 15 minutes per session for realism and correctness respectively, adjusted up from 10 minutes each after the first phase revealed higher average completion times.  These conditions were disclosed in the study information and consent
materials.  The application stored pseudonymous platform participant, study,
and session identifiers, structured judgments, written rationales, and
duration under server-side access controls.  The production-study materials below evidence the
participant-facing instructions and screenshots; they do not independently
evidence the recruitment screening, compensation, consent and withdrawal
process, researcher training, employer compliance, or IRB status reported
here.

\begin{figure*}[p]
\centering
\includegraphics[width=0.95\textwidth,trim={160bp 0 0 0},clip]{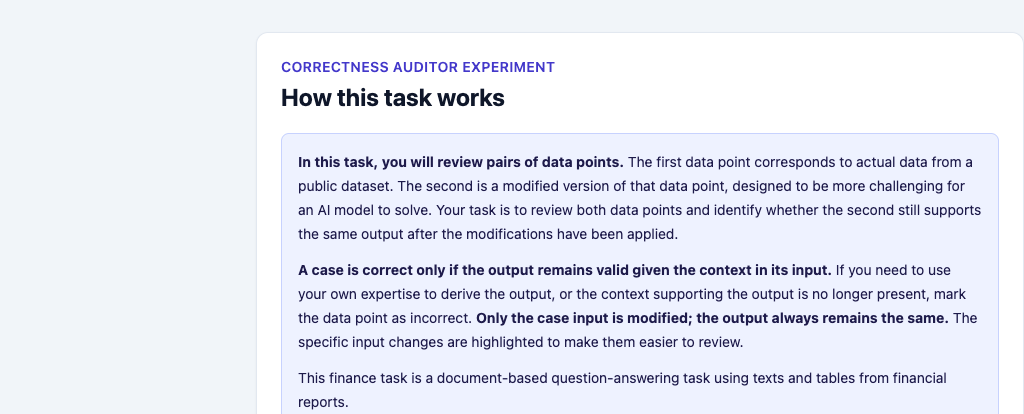}
\vspace{0.75em}

\includegraphics[width=0.95\textwidth,trim={160bp 0 0 0},clip]{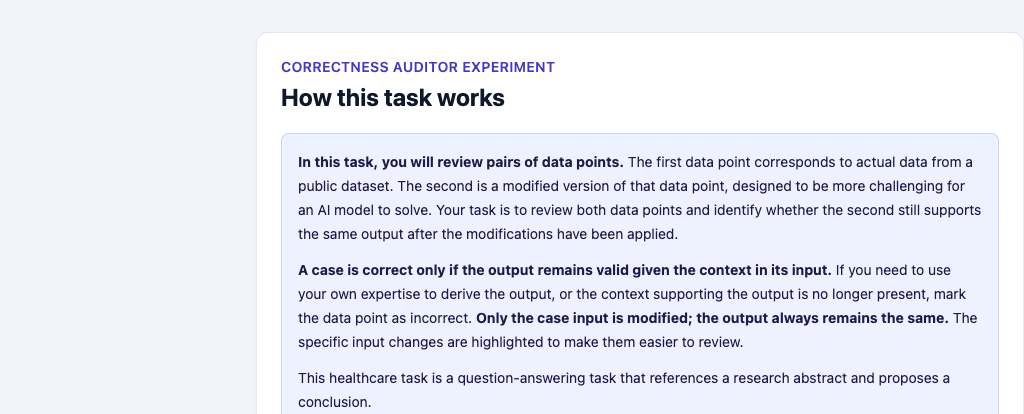}
\caption{Redacted correctness-study introduction screens for FinQA (top) and
PubMedQA (bottom).  The crops omit the embedded video preview and show no
participant or platform identifiers.}
\label{fig:participant-correctness-intros}
\end{figure*}

\begin{figure*}[p]
\centering
\includegraphics[width=0.95\textwidth,trim={100bp 0 100bp 0},clip]{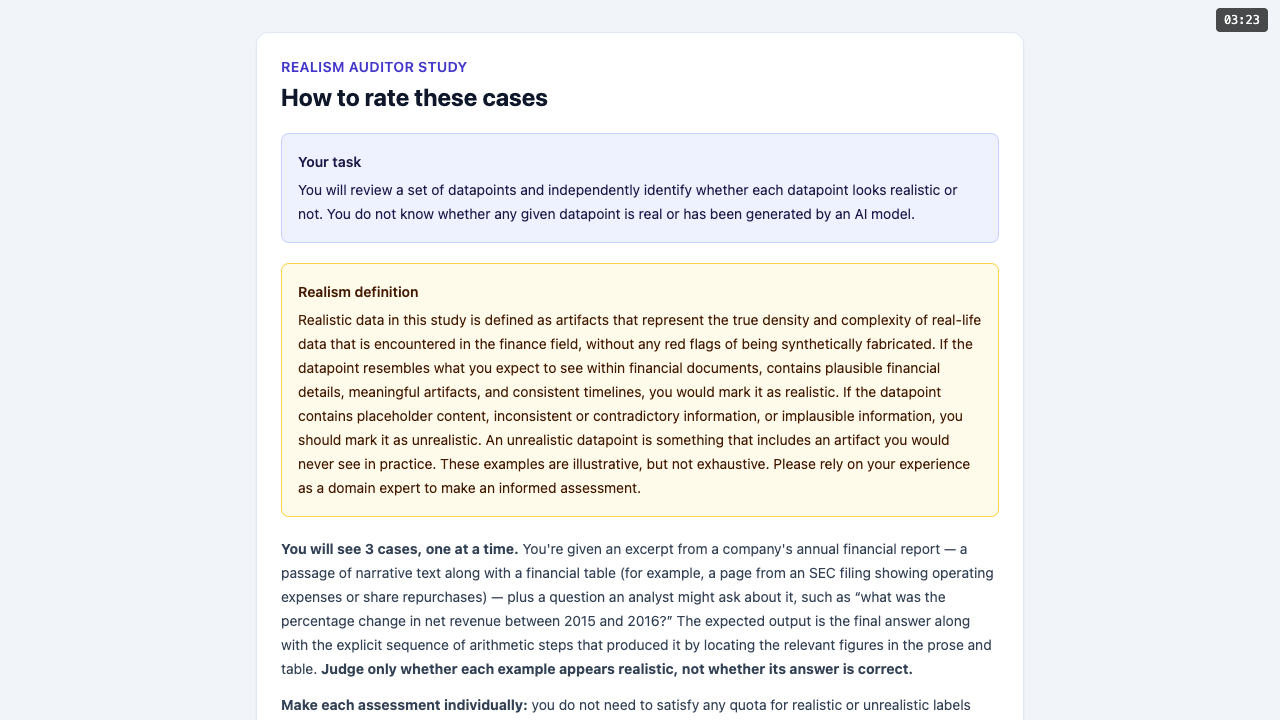}
\vspace{0.75em}

\includegraphics[width=0.95\textwidth,trim={100bp 0 100bp 0},clip]{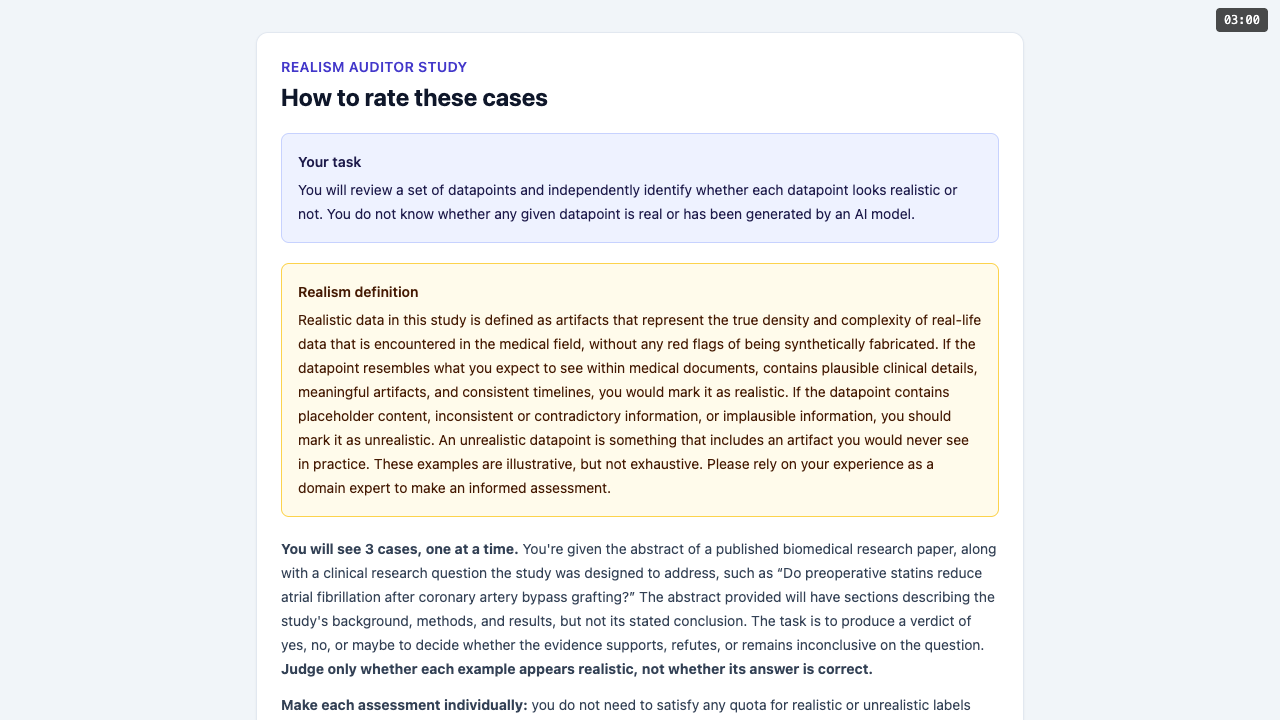}
\caption{Redacted realism-study introduction screens for FinQA (top) and
PubMedQA (bottom), including the benchmark-specific realism definitions and
task descriptions.  No participant or platform identifiers are shown.}
\label{fig:participant-realism-intros}
\end{figure*}

\begin{figure*}[p]
\centering
\includegraphics[width=0.82\textwidth]{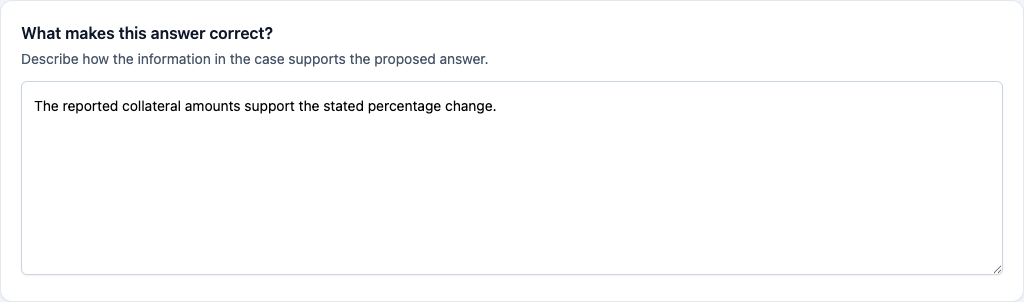}
\vspace{0.5em}

\includegraphics[width=0.82\textwidth]{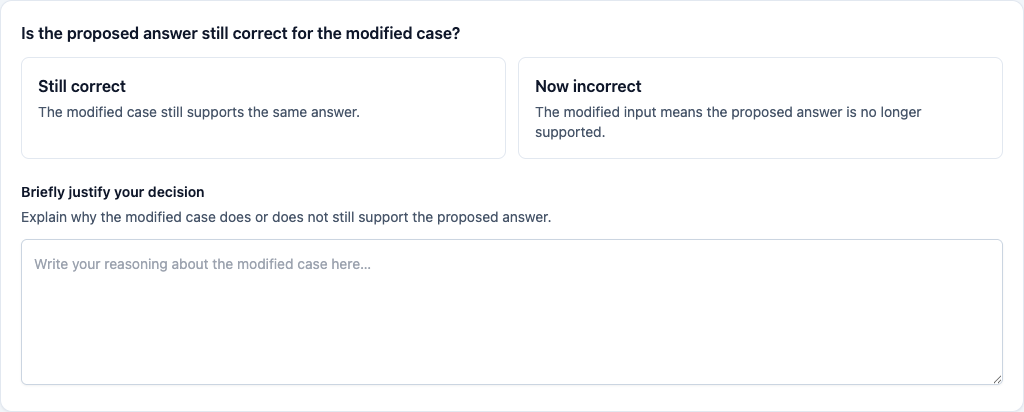}
\vspace{0.5em}

\includegraphics[width=0.82\textwidth]{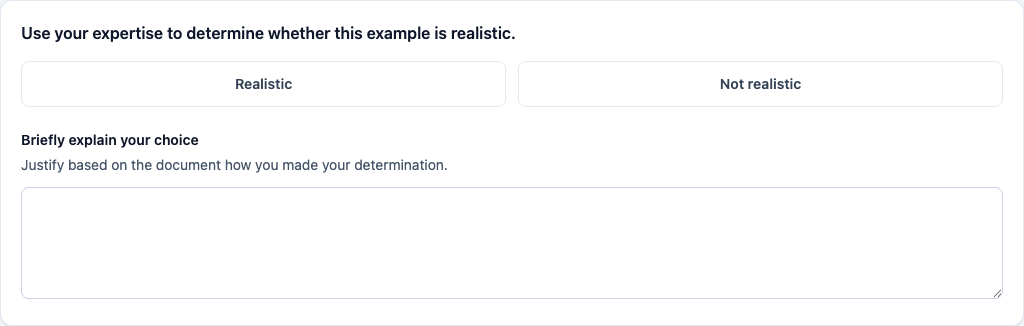}
\caption{Participant-facing response controls.  The original-case explanation
screen (top) is followed by the modified-case correctness judgment and
justification screen (middle); the realism judgment and rationale screen is
shown at bottom.  These controls are identical across FinQA and PubMedQA, so
only one capture of each is included.}
\label{fig:participant-response-controls}
\end{figure*}

\paragraph{Participant-facing interfaces.}
Participants had to finish a study-introduction video and acknowledge the
instructions before beginning.  Figures~\ref{fig:participant-correctness-intros}
and~\ref{fig:participant-realism-intros} reproduce the benchmark-specific
correctness and realism introductions.  The response controls are shown in
Figure~\ref{fig:participant-response-controls}.  The captures are redacted and
contain no participant, platform, or deployment identifiers.

\paragraph{Realism task.}
For each benchmark, the study contains 60 cases: 20 official benchmark cases,
20 synthetic cases authored by GPT-5.6 Sol, and 20 synthetic cases authored by
GPT-3.5 Turbo. Each task presents one case from each source in randomized
order.  Three participants label every case realistic or unrealistic and
provide rationales; the human label is the two-of-three majority.  We compare
these labels with the three-member feasibility ensemble on matched case banks.

\paragraph{Correctness task.}
Each task presents two original--mutated pairs drawn from real HARDEN
runs.  Participants first explain the original case's solution, having been informed of its correctness, then assess
whether the highlighted mutation preserves correctness and justify the
decision.  Each benchmark contributes 20 gold-labeled pairs, balanced between
ten correct and ten incorrect mutations, with three independent participant
judgments per pair.  The human label is again the participant majority.

\clearpage

\paragraph{Analysis.}
For realism, we report source-wise realistic rates in Table~\ref{tab:realism-source-rates}; direct check--human
agreement, Cohen's $\kappa$, Gwet's AC1, positive and negative agreement, and
$\phi$/MCC in Table~\ref{tab:realism-agreement}.  We additionally test the pre-specified official-versus-GPT-3.5
source comparison with a two-sided Fisher exact test.  For correctness, we
report confusion counts with correct as the positive class, accuracy,
specificity, $\kappa$, and an exact McNemar test with Holm adjustment across
the two benchmarks.

\begin{table*}[t]
\caption{Percentage of cases labeled realistic by the agentic check and the
two-of-three participant majority.  Each row contains 20 cases.}
\label{tab:realism-source-rates}
\centering
\small
\begin{tabular}{llrr}
\toprule
Benchmark & Source & Agentic check & Human majority \\
\midrule
FinQA & Official & 90\% & 60\% \\
FinQA & GPT-5.6 Sol & 95\% & 70\% \\
FinQA & GPT-3.5 Turbo & 0\% & 60\% \\
PubMedQA & Official & 100\% & 95\% \\
PubMedQA & GPT-5.6 Sol & 100\% & 100\% \\
PubMedQA & GPT-3.5 Turbo & 90\% & 60\% \\
\bottomrule
\end{tabular}
\end{table*}

\begin{table*}[t]
\caption{Realism agreement between the agentic check and participant majority
over 60 cases per benchmark.}
\label{tab:realism-agreement}
\centering
\scriptsize
\setlength{\tabcolsep}{4pt}
\begin{tabular}{lrrrrrr}
\toprule
Benchmark & Agreement & Cohen's $\kappa$ & Gwet's AC1 &
Positive agr. & Negative agr. & $\phi$/MCC \\
\midrule
FinQA & 51.7\% & $-0.031$ & 0.090 & 61.3\% & 35.6\% & $-0.031$ \\
PubMedQA & 85.0\% & 0.135 & 0.820 & 91.7\% & 18.2\% & 0.182 \\
\bottomrule
\end{tabular}
\end{table*}

\paragraph{Realism results.}
Tables~\ref{tab:realism-source-rates} and~\ref{tab:realism-agreement}
provide the detailed statistics behind the summary in
Section~\ref{sec:results}.  On FinQA, the check attains 95\% provenance
accuracy (38/40), but direct agreement with participant majorities is 51.7\%.
The observed participant acceptance rate is 60\% for both official and
GPT-3.5 cases ($p=1.000$, two-sided Fisher exact test). With 20 cases
per source, this estimate is imprecise and does not establish equal population
rates or imply that finance experts generally cannot distinguish the sources.

On PubMedQA, AC1 is 0.820, participant majorities separate official from
GPT-3.5 cases ($p=0.0197$), and feasibility check provenance accuracy is 55\% (22/40).
Together with the 85.0\% overall agreement and 18.2\% agreement regarding
unrealistic cases in
Table~\ref{tab:realism-agreement}, these results support the conclusion: the check is well aligned with humans on realistic examples but
over-permissive on unrealistic ones, motivating threshold recalibration.

\begin{table*}[t]
\caption{Correctness agreement on 20 gold-labeled mutated pairs per
benchmark.  Confusion counts are TP/FN/FP/TN with correct as positive.}
\label{tab:correctness-agreement}
\centering
\small
\begin{tabular}{llrrrrr}
\toprule
Benchmark & TP/FN/FP/TN & Accuracy & $\phi$/MCC &
Cohen's $\kappa$ & Specificity & McNemar Holm $p$ \\
\midrule
FinQA & 10/0/9/1 & 55.0\% & 0.229 & 0.100 & 10.0\% & 0.0039 \\
PubMedQA & 10/0/10/0 & 50.0\% & n/a & 0.000 & 0.0\% & 0.0039 \\
\bottomrule
\end{tabular}
\end{table*}

\paragraph{Correctness results.}
Table~\ref{tab:correctness-agreement} quantifies the low agreement summarized
in Section~\ref{sec:results}: participant specificity is 10.0\% on FinQA and
0.0\% on PubMedQA, with a significant correct-rate mismatch on both benchmarks
after Holm adjustment ($p=0.0039$). These results do not by
themselves determine whether strictness comes from superior check sensitivity
or an interface that makes semantic errors difficult for participants to
identify.  The secondary study below therefore evaluates the check directly
against a more informative labeled bank.

\subsection{Secondary correctness-check validation}
\label{app:secondary-correctness}

We therefore conduct a second validation against correctness labels assigned
by the co-authors rather than participant-majority labels.  Only pairs on which
the co-authors agree enter the evaluation set; split judgments are excluded
from all reported metrics, and the agreed co-author verdict serves as the gold label.  We report the completed FinQA
adjudication and leave a corresponding PubMedQA evaluation to future work.
The submitted FinQA bank contains 25 pairs: ten
manually mutated with the intent of making them incorrect, five manually
mutated while preserving correctness, five generated by a single-model
counterfactual operation, and five generated by a single-model label-flip
operation.  Four pairs receive split co-author judgments, leaving 21 for
evaluation.  The counterfactual prompt requests the smallest realistic change
that invalidates the original output; the label-flip prompt requests stronger
changes to decision-critical source evidence while preserving the question.
The production correctness check uses three independent Claude Opus~4.8 calls
at temperature zero and returns the two-of-three majority.

\begin{table*}[t]
\caption{Correctness-check agreement on the co-author-agreement evaluation set,
by construction method.  Numerators are matches to the agreed co-author
verdict; denominators exclude split judgments.}
\label{tab:secondary-correctness-strata}
\centering
\small
\begin{tabular}{lrrrr}
\toprule
Benchmark & Manual incorrect & Counterfactual & Label flip & Manual correct \\
\midrule
FinQA & 5/8 & 5/5 & 5/5 & 3/3 \\
\bottomrule
\end{tabular}
\end{table*}

\begin{table*}[h!]
\caption{Secondary correctness performance on the co-author-agreement
evaluation set, with correct as the positive class.  Confidence intervals are
Wilson 95\% intervals for accuracy.}
\label{tab:secondary-correctness-summary}
\centering
\small
\begin{tabular}{llrrr}
\toprule
Benchmark & TP/FN/FP/TN & Accuracy (95\% CI) &
Correct recall & Incorrect specificity \\
\midrule
FinQA & 3/1/2/15 & 85.7\% (65.4--95.0) & 75.0\% & 88.2\% \\
\bottomrule
\end{tabular}
\end{table*}

As summarized in Table~\ref{tab:secondary-correctness-summary}, the check
matches 18 of 21 agreed FinQA labels, with 81.6\% balanced accuracy,
Cohen's $\kappa=0.577$, and $\phi$/MCC${}=0.583$.
Table~\ref{tab:secondary-correctness-strata} shows that it rejects all ten
single-model counterfactual and label-flip examples and accepts all three
agreed manually degraded correct controls. These results show that the check
can distinguish explicit output-invalidating mutations from the tested valid
controls; the first study's low participant specificity should therefore not
be interpreted as evidence that the check merely rejects indiscriminately.

The three disagreements expose errors in both directions.  The check rejects
one manually constructed pair that both co-authors judge correct, treating a
changed table header as decision-critical despite compensating narrative
context.  It accepts two pairs that both co-authors judge incorrect, overlooking
a location mismatch when the numeric operands remain and a shifted temporal
bucket in a financial table.  All three feasibility check decisions are unanimous.

This secondary validation is deliberately small and class-imbalanced, and four
of the 25 submitted FinQA pairs are excluded after split judgments.  Its
model-generated negatives were prompted with the original output and may be
more explicit than naturally occurring invalid mutations; the three agreed
correct controls are also all manually authored.  We therefore interpret the
results as evidence that the check is a useful but incomplete validity filter:
it reliably detects the tested model-generated changes and preserves the
agreed controls, but can both over-reject benign inconsistencies and
over-accept subtler errors involving entity alignment or temporal semantics.
Larger adjudicated studies, including a corresponding PubMedQA evaluation,
are needed before treating a check acceptance as conclusive proof of validity.







\end{document}